\pdfoutput=1

\documentclass[11pt]{article}

\usepackage[preprint]{acl}

\usepackage{times}
\usepackage{latexsym}

\usepackage[T1]{fontenc}

\usepackage[utf8]{inputenc}

\usepackage{microtype}

\usepackage{inconsolata}

\usepackage{graphicx}

\usepackage{xcolor} 

\usepackage{amsthm,amssymb}
\usepackage{amsmath}
\usepackage{mathrsfs}
\usepackage{multirow,multicol,booktabs}
\usepackage{enumitem} 
\usepackage{makecell} 
\usepackage{algorithm}
\usepackage{algpseudocode}
\usepackage{colortbl}
\definecolor{lightblue}{RGB}{203, 228, 253}
\newcommand{\blue}{\cellcolor{lightblue}}

\title{Better Process Supervision with Bi-directional Rewarding Signals}

\author{
    Wenxiang Chen$^{1}$\thanks{{ }\ Equal contributions. $^\dagger$ Correspondence to: \texttt{chenwx23@\\m.fudan.edu.cn, \{tgui, qz\}@fudan.edu.cn}}, \ \  Wei He$^{1*}$, \ \ Zhiheng Xi$^{1*}$, \\
    \textbf{Honglin Guo}$^1$\textbf{,} \ \ \textbf{Boyang Hong}$^{1}$\textbf{,} \ \ \textbf{Jiazheng Zhang}$^1$\textbf{,} \ \ \textbf{Rui Zheng}$^1$\textbf{,} \\
    \textbf{Nijun Li}$^2$\textbf{,} \ \ \textbf{Tao Gui}$^{3\dagger}$\textbf{,} \ \ \textbf{Yun Li}$^{2}$\textbf{,} \ \ \textbf{Qi Zhang}$^{1\dagger}$\textbf{,} \ \ \textbf{Xuanjing Huang}$^1$ \\
    \normalsize{$^1$  School of Computer Science,\ Fudan University} \\
    \normalsize{$^2$ Cognitive AI Lab, Shanghai Huawei Technologies, \ China} \\
    \normalsize{$^3$  Institute of Modern Languages and Linguistics,\ Fudan University} \\
}

\begin{document}
\maketitle
\begin{abstract}

Process supervision, i.e., evaluating each step, is critical for complex large language model (LLM) reasoning and test-time searching with increased inference compute. 
Existing approaches, represented by process reward models (PRMs), primarily focus on rewarding signals up to the current step, exhibiting a one-directional nature and lacking a mechanism to model the distance to the final target.
To address this problem, we draw inspiration from the A* algorithm, which states that an effective supervisory signal should simultaneously consider the incurred cost and the estimated cost for reaching the target. Building on this key insight, we introduce \textbf{BiRM}, a novel process supervision model that not only evaluates the correctness of previous steps but also models the probability of future success. 
We conduct extensive experiments on mathematical reasoning tasks and demonstrate that BiRM provides more precise evaluations of LLM reasoning steps, achieving an improvement of $3.1\%$ on Gaokao2023 over PRM under the Best-of-N sampling method. Besides, in search-based strategies, BiRM provides more comprehensive guidance and outperforms ORM by $5.0\%$ and PRM by $3.8\%$ respectively on MATH-500\footnote{\phantom{} Our code and data are available at: \url{https://github.com/chenwxOggai/BiRM}.}.

\end{abstract}

\section{Introduction}
With the rapid development of LLMs, how to supervise them has become a key research challenge, especially for complex tasks like long-term reasoning \cite{DBLP:conf/nips/ZelikmanWMG22, openai-o1, DBLP:conf/icml/WanFWM00024}. Previous work has explored training process supervision models to provide dense supervision on each step \cite{DBLP:journals/corr/abs-2211-14275, DBLP:conf/iclr/LightmanKBEBLLS24, DBLP:conf/acl/WangLSXDLCWS24}, which is intuitively and practically better than outcome supervision models \cite{DBLP:journals/corr/abs-2110-14168} that only provide sparse signals on the final answer. During test-time, process supervision models can further guide the search of LLMs or perform solution re-ranking by allocating more inference compute \cite{DBLP:journals/corr/abs-2408-03314,DBLP:journals/corr/abs-2407-21787, wu2024inference}.

However, existing approaches, represented by process reward models (PRMs) from OpenAI \cite{DBLP:conf/iclr/LightmanKBEBLLS24}, typically focus on providing one-directional reward signals on the reasoning steps that have already been generated, without consciously considering the probability of future success \cite{DBLP:conf/naacl/YuGW24, zhang2025lessons}. Specifically, while they can accurately distinguish between correct and incorrect steps at the current state (i.e., backward supervision), their ability to identify which partial solution is most likely to reach the correct final answer (i.e., forward supervision) is not guaranteed, leading to sub-optimal performance in guiding effective next-step reasoning \cite{stroebl2024inference, wang2025examining}.

\begin{figure}[t]
    \centering
    \includegraphics[width=\linewidth]{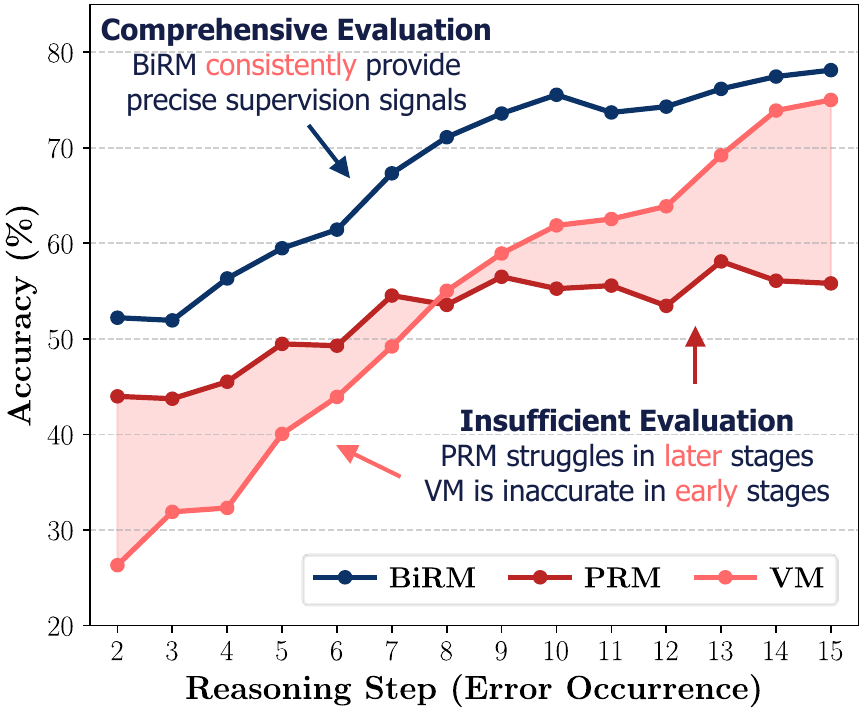}
    \caption{Error-detection accuracy across different steps, where step 1 and steps beyond 15 are truncated for better visualization. We evaluate the process reward model (PRM), value model (VM), and \textbf{BiRM} on PRMBench.}
    \vspace{-5pt}
    \label{fig:PRMbench}
\end{figure}

To address this challenge, we draw inspiration from the classic A* algorithm, and introduce \textbf{BiRM}, a novel process supervision model that provides bidirectional rewarding signals. Classically, the A* algorithm \cite{DBLP:journals/tssc/HartNR68} states that an appropriate supervisory signal should take two aspects into account: the cumulative cost up to the current step, and the estimated probability of reaching the target \cite{DBLP:conf/iclr/ZhuangC0MBRS024, DBLP:journals/corr/abs-2406-14283}. Motivated by this key insight, we redesign the process supervision signals, which should not only assess the correctness of steps taken so far, but also evaluate the future success probability of the partial solution. Specifically, BiRM introduces a value model (VM) head to help model the forward supervision signal \cite{DBLP:conf/naacl/YuGW24, DBLP:journals/corr/abs-2408-11791}, so that it can estimate both the correctness and success probability of a reasoning prefix/partial solution (Section \ref{sec:4}).

To validate our motivation, we conduct a preliminary analysis on \textsc{PRMBench} \cite{song2025prmbench}, a benchmark designed to evaluate the capability of process supervision models. We include PRM and VM as baselines, where the former estimates the correctness of partial solutions, and the latter estimates the future success probability. As shown in Figure \ref{fig:PRMbench}, PRM performs better at detecting error steps in the early stages of reasoning, while VM performs better in the later stages. This indicates that each baseline has limitations, which aligns with the intuition we derive from the A* algorithm. In contrast, BiRM outperforms both of them in all stages, demonstrating the comprehensiveness and effectiveness of our approach.

We then perform extensive experiments on three mathematical reasoning tasks: GSM8K, MATH-500 and Gaokao2023 \cite{DBLP:conf/acl/Liao0LW024} to demonstrate the effectiveness of BiRM across different model series and search strategies. For example, BiRM trained on Qwen2.5-7B-Base achieves a $3.1\%$ improvement on Gaokao2023 over PRM using Best-of-N sampling. Additionally, in beam search with a total sampling size of $100$, BiRM further surpasses PRM by $3.8\%$ and ORM by $5.0\%$.

In summary, our contributions are as follows:

\begin{itemize}[leftmargin=*]
    \item We draw inspiration from A* algorithm and propose BiRM, a novel process supervision model that provides bidirectional rewarding signals. 
    \item  We conduct extensive experiments on math reasoning tasks to demonstrate its effectiveness in solution re-ranking and trajectory searching.
    \item We present an in-depth analysis and demonstrate that BiRM is orthogonal to existing open-source supervision models, highlighting its robustness and generalization capabilities.
\end{itemize}

\section{Related Work}
\subsection{Enhancing Mathematical Reasoning Capabilities of LLMs}
Mathematical reasoning tasks remain a significant challenge for LLMs \cite{gpt4o, DBLP:journals/corr/abs-2408-03314}. Researchers have conducted extensive studies on both train-time and test-time improvements. At train-time, supervised fine-tuning is a well-established approach. Its core idea is to construct large-scale, high-quality datasets to enhance performance \cite{DBLP:conf/acl/Liao0LW024, DBLP:conf/iclr/YuJSYLZKLWL24, DBLP:conf/nips/TongZWWH24}.
On the other hand, experimental results from \texttt{Openai-o1} \cite{openai-o1} and \texttt{DeepSeek-R1} \cite{deepseekai2025deepseekr1incentivizingreasoningcapability} highlight the promising potential of test-time scaling laws. Vanilla sampling methods like Best-of-N sampling \cite{liu2025pairwise} and search-based strategies such as beam search, A*, and MCTS \cite{DBLP:conf/iclr/ZhuangC0MBRS024, DBLP:conf/icml/WanFWM00024, DBLP:journals/corr/abs-2410-02884} have all achieved remarkable performance by allocating more computational resources at test-time. In this work, we focus on improving LLM's performance during the test-time phase.

\subsection{Process Supervision Models in LLM Reasoning}
LLMs can leverage an additional supervision model to achieve accurate test-time reasoning. Mainstream approaches can be divided into outcome reward models (ORMs) and process reward models (PRMs). ORMs are trained with rule-based labeled data and assign one score to the entire solution path \cite{DBLP:journals/corr/abs-2110-14168, DBLP:conf/naacl/YuGW24}. This method achieves striking results in reasoning models like Deepseek-R1 but struggles with other tasks where the answers are highly open-ended. On the other hand, PRMs evaluate each intermediate steps in the trajectory, providing more granular reward signals \cite{DBLP:conf/iclr/LightmanKBEBLLS24, DBLP:journals/corr/abs-2211-14275, zhang2025lessons}. Depending on the practical implementation, there are several variants of PRMs: \textit{(1)} \textit{Value Models} (VMs, \citealp{DBLP:conf/acl/WangLSXDLCWS24, DBLP:journals/corr/abs-2406-06592}) use Monte Carlo estimation to label steps, reducing the burden of manual annotation. The resulting labels represent the probability of future success, essentially making PRMs a type of value model. \textit{(2)} \textit{Generative Reward Models} \cite{DBLP:journals/corr/abs-2408-15240} leverage the text generation capabilities of LLMs, providing natural language feedback, rather than traditional numerical scores.

\section{Motivation}
\subsection{Task Formulation}
Given a mathematical question $ q $, a large language model $ \pi $ generates a sequence of reasoning steps to solve the problem. The complete reasoning trajectory, i.e., chain-of-thought \cite{DBLP:conf/nips/Wei0SBIXCLZ22}, can be denoted as $ \tau = \{ s_1, s_2, \dots, s_m \} $, where $ s_i $ represents the $ i $-th step and $ m $ is the number of total reasoning steps. 

\subsection{The Limitations of PRMs}
\begin{figure*}[t]
    \centering
    \includegraphics[width=0.88\textwidth]{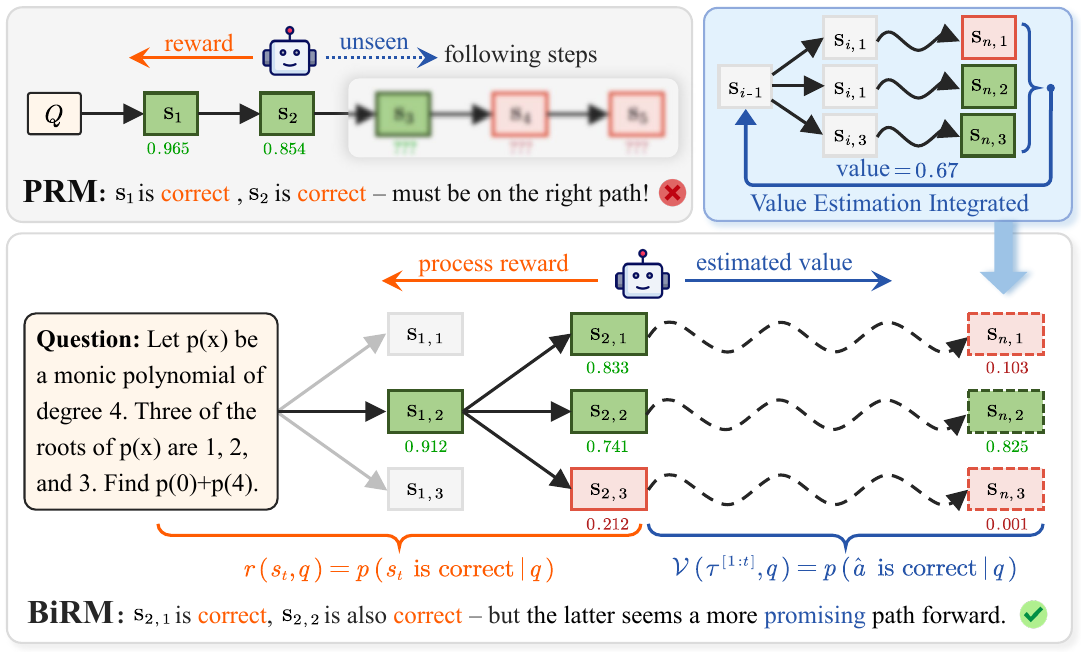}
    \caption{An example of our proposed \textbf{BiRM} compared with traitional Process Reward Models (PRMs). Given a question $q$, PRMs only consider the accumulated rewards up to the current step. In contrast, BiRM takes into account two aspects: the correctness rewards received so far and the probability of reaching correct final answers.}
    \vspace{-5pt}
    \label{fig:Main}
\end{figure*}

PRMs are typically trained to assign a numerical score to each intermediate reasoning step, evaluating their correctness. For a partial trajectory $ \tau^{[1:t]} = \{ s_1, s_2, \dots, s_t \} $, PRM can provide an reward score for step $ s_i $:
\begin{equation}\label{eq:PRM-single-reward}
r(s_i,q) = p(s_i \text{ is correct} \mid q),
\end{equation}
where $ r(\cdot) $ represents the process-based reward function provided by PRM. Further, the correctness of the partial trajectory $ \tau^{[1:t]} $ can be expressed as the accumulative correctness reward of all intermediate steps, following \citet{DBLP:conf/iclr/LightmanKBEBLLS24}:
\begin{equation}
\begin{aligned}
\mathcal{R}(\tau^{[1:t]},q) &= p([s_1, s_2, \dots, s_t] \text{ is correct} \mid q) \\
&= \prod_{i=1}^{t} p(s_i \text{ is correct} \mid q) = \prod_{i=1}^{t} r(s_i,q) \nonumber.
\end{aligned}
\end{equation}
This equation highlights the one-directional scoring nature of PRMs, which evaluate whether the sampled trajectory $ \{ s_1, s_2, \dots, s_t \} $ is correct given the problem $ q $. Instead, for the potential future paths $ \{ s_{t+1}, s_{t+2}, \dots, s_m \} $ starting from the current state $ s_t $, PRMs lack the capability to provide effective guidance, as Figure \ref{fig:Main} illustrates.

\subsection{Inspiration from the A* Search Algorithm}
To address this limitation, we draw inspiration from the A* algorithm. Originally, A* is a heuristic graph search algorithm designed to find the optimal path \cite{DBLP:journals/tssc/HartNR68}. The key insight from A* is that a good supervision signal should simultaneously consider two aspects: the accumulative cost $g(n)$ up to the current step and the future cost $h(n)$ to the target. The final value of a step is given by $f(n) = g(n) + h(n)$.

In the context of LLM mathematical reasoning, we argue that a good supervision signal should not only consider the correctness of previous steps (i.e., backward supervision) but also model the probability of future success (i.e., forward supervision). On the one hand, PRM can naturally function as $g(\cdot)$. In other words, PRM can use its one-directional scoring ability to provide rewards for the partial solution up to the current step $ s_t $:
\begin{equation}
g(s_t) = \text{Agg}(r(s_1), r(s_2), \dots, r(s_t)) = \mathcal{R}(\tau^{[1:t]}) \nonumber,
\end{equation}
where $ \text{Agg} \in \{ \prod, \min, \max, \text{avg} \} $ stands for an aggregation function to summarize the accumulative rewards of all steps from $s_1$ to $s_t$.

On the other hand, to heuristically model the probability of reaching the correct final answer, we seek to utilize a value model (VM) to play the role of $h(\cdot)$. For the partial solution $ \tau^{[1:t]} $, a forward-looking VM can provide a reliable probability estimation:
\begin{equation}\label{eq:PRM-single-value}
\begin{aligned}
    h(s_t) &= \mathcal{V}(\tau^{[1:t]}, q) \\
    & = \mathbb{E}_{\hat{a} \sim \pi(\cdot \mid \tau^{[1:t]}, q)} \left[ p(\hat{a} \text{ is correct} \mid q) \right].
\end{aligned}
\end{equation}
Here, $\hat{a}$ represents the final answer predicted by the LLM $ \pi $, and $\mathcal{V}(\cdot)$ denotes the estimtation of VM for whether the partial trajectory can reach the correct answer. In practical implementations, the VM and PRM share the same model architecture, but differ in the meaning of training labels, which fundamentally trains the VM as a reliable predictive estimator. We will discuss more details in Section \ref{sec:4.2}. Finally, the complete value function can be expressed as:
\begin{equation}\label{eq:BiRM-eval}
f(s_t) = g(s_t) + \beta \cdot h(s_t),
\end{equation}
where the coefficient $\beta$ balances the importance of the $g(s_t)$ and $h(s_t)$ terms. When a step $s_i$ has a higher $f(s_i)$ value, it indicates that this step is more promising among multiple candidates, thus contributing to more effective next-step reasoning.

\section{BiRM, a Bidirectional Process Supervision Model}\label{sec:4}
\subsection{Training Methodology}
For a query $ q $ from the training question set $ \mathcal{Q} $, we first sample $ N $ solutions from the generator $ \pi $. Then, we annotate each intermediate step of these solutions, i.e., annotating step-level labels.  The resulting dataset $ \mathcal{D} $ for query $ q $ can be formalized as $ \mathcal{D}_q = \{ \tau_i, \{ y_i^1, y_i^2, \dots, y_i^j, \dots \} \}_{i=1}^{N} $, where $ \tau_i $ denotes the $ i $-th sampled trajectory, and $ y_i^j $ represents the step label for the $ j $-th step in the $ i $-th solution. We will introduce more annotation strategies in Section \ref{sec:4.2}.

Following \citet{DBLP:conf/naacl/YuGW24}, we implement the vanilla PRM by adding a linear layer for reward prediction after the last hidden layer of the LLM. We also retain the original language modeling head. Formally, a vanilla PRM $\mathcal{R} ({\theta, \phi_R}) $ is parameterized by base model parameters $ \theta $ and reward head parameters $ \phi_R $. The training objective of PRM is to minimize the mean squared error (MSE) loss between the predicted reward scores and the binary step-level reward labels. Thus, we have:
\begin{equation}
\begin{aligned}
    &\mathcal{L}_{\text{PRM}}(\theta, \phi_R) \\
    &= \frac{1}{|\mathcal{Q}|}\sum_{q\in \mathcal{Q}} \left[ \mathbb{E}_{\tau\sim \pi(\cdot|q)}  \sum_{t=1}^{m} \big(\hat{r}_{\theta, \phi_R} (s_t,q) - r^t\big)^2 \right] \nonumber,
\end{aligned}
\end{equation}
where $ \hat{r}(s_t,q) $ represents the predicted reward score for the $ t $-th step (Equation \ref{eq:PRM-single-reward}), and $ r^t $ denotes the ground truth step label. $ m $ represents the total number of steps in solution $ \tau$.

Furthermore, to alleviate the one-directional limitation of PRM, we introduce an additional value head to guide process supervision. Specifically, BiRM $ \mathcal{M}({\theta, \phi_R, \phi_V})$ is parameterized by three components: $ \theta $ represents the base model parameters, $ \phi_R $ represents the reward head, and $ \phi_V $ corresponds to the value head. The overall training objective of BiRM is to jointly minimize the discrepancy between the predicted reward score and the reward label, as well as between the value score and the value label. 
Similar to the vanilla PRM, we employ MSE loss for the value head:
\begin{equation}
\begin{aligned}
    &\mathcal{L}_{\text{VM}}(\theta, \phi_V) = \frac{1}{|\mathcal{Q}|}\sum_{q\in \mathcal{Q}} \\
    & \qquad \left[ \mathbb{E}_{\tau\sim \pi(\cdot|q)} \sum_{t=1}^{m} \big(\hat{\mathcal{M}}_{\theta, \phi_V} (\tau^{[1:t]}, q) - v^t\big)^2 \right] \nonumber,
\end{aligned}
\end{equation}

where $ \hat{\mathcal{M}}_{\theta, \phi_V} $ represents the estimated success probability for the partial solution $ \tau^{[1:t]} $ (Equation~\ref{eq:PRM-single-value}), and $ v^t $ denotes the value label for $ s_t $.

In this way, the optimized BiRM considers not only the actual accumulative rewards obtained so far, but also the potential of reaching correct final answers (Figure \ref{fig:Main}). The complete loss function for BiRM can be defined as:
\begin{equation}\label{eq:birm-loss}
\mathcal{L}_{\text{BiRM}}(\theta, \phi_R, \phi_V) = \mathcal{L}_{\text{PRM}}(\theta, \phi_R) + c \cdot \mathcal{L}_{\text{VM}}(\theta, \phi_V).
\end{equation}
We use a coefficient $ c $ to balance the importance of the reward term $ \mathcal{L}_{\text{PRM}} $ and the value term $ \mathcal{L}_{\text{VM}} $. 

\subsection{Step Label Annotation Strategies}
\label{sec:4.2}
In this section, we discuss our annotation strategies for two kinds of BiRM training labels.

\paragraph{Reward Labels.}
Reward labels are defined as the correctness of each current step, represented as a binary label. We use the MetaMath dataset \cite{DBLP:conf/iclr/YuJSYLZKLWL24} as our training data. We first perform supervised fine-tuning on the base model to obtain the generators. Then, we sample $15$ rollouts for each query and use \texttt{Deepseek-V3} \cite{DBLP:journals/corr/abs-2412-19437} to annotate the correctness of each step. Detailed annotation procedures and prompts are provided in Appendix \ref{appendix:reward-label}.

\paragraph{Value Labels.}\label{sec:value-label}
A key challenge in implementing the value head is to accurately estimate value labels for the partial solution $\tau^{[1:t]}$. We employ multiple strategies to address this problem.

\textbf{MC-based estimation} \cite{DBLP:conf/acl/WangLSXDLCWS24} is a widely used method for automated labeling, which can be categorized into soft-label and hard-label annotations. Specifically, we sample $N$ rollouts from an intermediate step in the trajectory. If $M$ of them are correct, the soft-label for the current step can be defined as: $\text{label}(s_t) = \frac{M}{N}$. In contrast, the hard-label method suggests that if any of the rollouts reaches the target, then $\text{label}(s_t) = 1$.

The essence of Monte Carlo estimation is to assess the potential of reaching correct final answer from the current step and assign this probability to the step label. Thus, for estimating a partial trajectory, we can formally express it as:
\begin{equation}
\mathcal{V}(\tau^{[1:t]}, q) \approx \frac{1}{N} \sum_{i=1}^{N} \mathbb{I}(\hat{a}_i \text{ is correct} \mid \tau^{[1:t]}, q) \nonumber.
\end{equation}
As the number of rollouts $N$ increases, the estimated value label becomes more accurate. Following \citet{DBLP:conf/acl/WangLSXDLCWS24}, we sample $8$ solutions for each intermediate step and analyze the effectiveness of both soft-label and hard-label approaches.

\textbf{Outcome-supervised estimation} \cite{DBLP:conf/naacl/YuGW24} states that using the oucome label alone is sufficient to provide probability estimatation for each reasoning steps. The underlying idea is that during the training phase, we can replicate the final answer’s correctness label across all intermediate steps. The resulting value model implicitly learns to foresee the future, predicting potential final outcome (i.e. value) for partial solutions. Compared to MC estimation, outcome-supervised estimation has higher data efficiency, but the shortcoming is that the automatically learned estimation in this way is less accurate.

\section{Experiments}
\begin{table*}[t]
    \centering
    \resizebox{\textwidth}{!}{
    \begin{tabular}{@{}ll|c|ccccccccc@{}}
    \toprule
    \multirow{2}{*}{\textbf{Models}} & \multirow{2}{*}{\textbf{Methods}} & \multirow{2}{*}{\bf Avg.} & \multicolumn{3}{c|}{\textbf{GSM8K}} & \multicolumn{3}{c|}{\textbf{MATH-500}} & \multicolumn{3}{c}{\textbf{Gaokao2023}}  \\
    
    \cmidrule(lr){4-12}
    
    & & & @\textit{128} & @\textit{256} & \multicolumn{1}{c|}{@\textit{512}} & @\textit{128} & @\textit{256} & \multicolumn{1}{c|}{@\textit{512}} & @\textit{128} & @\textit{256} & \multicolumn{1}{c}{@\textit{512}} \\ 
    \midrule

    \multirow{7}{*}{\textbf{Qwen2.5-3B}}
    & Greedy & $46.8$ & \multicolumn{3}{c|}{---------\; $73.1$ \;---------} & \multicolumn{3}{c|}{---------\; $40.2$ \;---------} & \multicolumn{3}{c}{---------\; $27.0$ \;---------} \\ 
    & Majority Vote & $58.1$ & $85.1$ & $85.0$ & \multicolumn{1}{c|}{$85.3$} & $52.5$& $53.0$ & \multicolumn{1}{c|}{$53.8$} & $35.8$ & $36.3$ & \multicolumn{1}{c}{$36.1$} \\
    & ORM & $58.9$ & $88.1$ & $88.1$ & \multicolumn{1}{c|}{$88.1$} & $52.1$ & $51.8$ & \multicolumn{1}{c|}{$52.2$} & $37.2$ & $37.0$ & \multicolumn{1}{c}{$35.8$} \\
    & PRM & $59.9$ & $\mathbf{88.5}$ & $88.3$ & \multicolumn{1}{c|}{$88.0$} & $54.6$ & $54.1$ & \multicolumn{1}{c|}{$54.2$} & $\mathbf{37.3}$ & $37.1$ & \multicolumn{1}{c}{$37.2$} \\
    & ER-PRM & $58.8$ & $88.0$ & $88.0$ & \multicolumn{1}{c|}{$87.7$} & $52.6$ & $52.3$ & \multicolumn{1}{c|}{$52.0$} & $36.2$ & $36.3$ & \multicolumn{1}{c}{$35.8$} \\
    & Math-Shepherd & $59.0$ & $87.3$ & $87.2$ & \multicolumn{1}{c|}{$87.0$} & $53.2$ & $53.4$ & \multicolumn{1}{c|}{$53.8$} & $36.6$ & $36.4$ & \multicolumn{1}{c}{$36.1$} \\
    & \blue{\textbf{BiRM}} & \blue{$\mathbf{61.0}$} & \blue{$88.4$} & \blue{$\mathbf{88.6}$} & \multicolumn{1}{c|}{\blue{$\mathbf{88.9}$}} & \blue{$\mathbf{55.9}$} & \blue{$\mathbf{56.1}$} & \multicolumn{1}{c|}{\blue{$\mathbf{57.4}$}} & \blue{$36.9$} & \blue{$\mathbf{37.8}$} & \multicolumn{1}{c}{\blue{$\mathbf{38.7}$}} \\
    \midrule

    \multirow{7}{*}{\textbf{Qwen2.5-7B}}
    & Greedy & $52.3$ & \multicolumn{3}{c|}{---------\; $78.5$ \;---------} & \multicolumn{3}{c|}{---------\; $45.0$ \;---------} & \multicolumn{3}{c}{---------\; $33.5$ \;---------} \\ 
    & Majority Vote & $63.6$ & $88.1$ & $88.0$ & \multicolumn{1}{c|}{$87.8$} & $57.3$ & $57.5$ & \multicolumn{1}{c|}{$57.6$} & $45.5$ & $45.4$ & \multicolumn{1}{c}{$45.2$} \\
    & ORM & $64.7$ & $92.0$ & $91.6$ & \multicolumn{1}{c|}{$91.3$} & $59.6$ & $59.9$ & \multicolumn{1}{c|}{$59.4$} & $43.6$ & $43.5$ & \multicolumn{1}{c}{$41.3$} \\
    & PRM & $66.3$ & $92.7$ & $92.8$ & \multicolumn{1}{c|}{$92.9$} & $60.3$ & $60.1$ & \multicolumn{1}{c|}{$58.4$} & $45.8$ & $46.2$ & \multicolumn{1}{c}{$47.3$} \\
    & ER-PRM & $66.2$ & $92.2$ & $92.1$ & \multicolumn{1}{c|}{$92.2$} & $59.7$ & $59.2$ & \multicolumn{1}{c|}{$59.0$} & $47.0$ & $47.2$ & \multicolumn{1}{c}{$47.3$} \\
    & Math-Shepherd & $66.3$ & $92.1$ & $92.2$ & \multicolumn{1}{c|}{$91.7$} & $60.3$ & $60.2$ & \multicolumn{1}{c|}{$60.4$} & $46.4$ & $47.0$ & \multicolumn{1}{c}{$46.5$} \\
    & \blue{\textbf{BiRM}} & \blue{$\mathbf{68.3}$} & \blue{$\mathbf{93.1}$} & \blue{$\mathbf{93.3}$} & \multicolumn{1}{c|}{\blue{$\mathbf{93.2}$}} & \blue{$\mathbf{62.4}$} & \blue{$\mathbf{62.3}$} & \multicolumn{1}{c|}{\blue{$\mathbf{63.4}$}} & \blue{$\mathbf{47.7}$} & \blue{$\mathbf{49.1}$} & \multicolumn{1}{c}{\blue{$\mathbf{50.4}$}} \\
    \midrule

    \multirow{7}{*}{\textbf{Llama3.1-8B}}
    & Greedy & $34.7$ & \multicolumn{3}{c|}{---------\; $55.7$ \;---------} & \multicolumn{3}{c|}{---------\; $31.2$ \;---------} & \multicolumn{3}{c}{---------\; $17.1$ \;---------} \\ 
    & Majority Vote & $46.4$ & $72.1$ & $72.0$ & \multicolumn{1}{c|}{$72.3$} & $39.2$ & $40.2$ & \multicolumn{1}{c|}{$41.1$} & $26.5$ & $27.2$ & \multicolumn{1}{c}{$27.1$} \\ 
    & ORM & $50.3$ & $84.1$ & $84.5$ & \multicolumn{1}{c|}{$85.0$} & $41.5$ & $40.9$ & \multicolumn{1}{c|}{$40.8$} & $25.4$ & $25.2$ & \multicolumn{1}{c}{$24.9$} \\ 
    & PRM & $51.5$ & $84.1$ & $84.8$ & \multicolumn{1}{c|}{$85.2$} & $42.5$ & $42.2$ & \multicolumn{1}{c|}{$41.8$} & $28.2$ & $27.7$ & \multicolumn{1}{c}{$27.3$} \\ 
    & ER-PRM & $50.6$ & $84.8$ & $85.3$ & \multicolumn{1}{c|}{$85.8$} & $41.3$ & $41.0$ & \multicolumn{1}{c|}{$40.2$} & $25.7$ & $26.1$ & \multicolumn{1}{c}{$24.9$} \\
    & Math-Shepherd & $51.3$ & $84.4$ & $84.9$ & \multicolumn{1}{c|}{$85.3$} & $42.7$ & $42.9$ & \multicolumn{1}{c|}{$43.6$} & $25.8$ & $25.8$ & \multicolumn{1}{c}{$26.2$} \\
    & \blue{\textbf{BiRM}} & \blue{$\mathbf{54.1}$} & \blue{$\mathbf{86.1}$} & \blue{$\mathbf{87.2}$} & \multicolumn{1}{c|}{\blue{$\mathbf{87.8}$}} & \blue{$\mathbf{45.4}$} & \blue{$\mathbf{45.4}$} & \multicolumn{1}{c|}{\blue{$\mathbf{45.6}$}} & \blue{$\mathbf{29.4}$} & \blue{$\mathbf{30.0}$} & \multicolumn{1}{c}{\blue{$\mathbf{29.6}$}} \\
    \midrule
    \end{tabular}
    }
    \caption{Performance of Best-of-N sampling on GSM8K, MATH-500 and Gaokao2023 with three base models. The accuracy of the BoN solution is utilized as the evaluation metric. The results are reported as the average accuracy across five random seeds. 
    @\textit{128}, @\textit{256}, and @\textit{512} denote the accuracy with Best-of-128, Best-of-256, and Best-of-512 sampling, respectively. The results of greedy decoding are independent of $N$ and are listed for comparison purposes. The best results are marked in \textbf{bold}. 
    }
    \label{tab:main-result}
\end{table*}
\subsection{Experimental Setup}
\paragraph{Tasks.}
We conduct experiments using three widely used math reasoning datasets: GSM8K \cite{DBLP:journals/corr/abs-2110-14168}, MATH-500 \cite{DBLP:conf/iclr/LightmanKBEBLLS24}, and an out-of-domain (OOD) dataset Gaokao2023 \cite{DBLP:conf/acl/Liao0LW024} to evaluate the generalization ability of BiRM. Besides, we test our method on three base models across different model sizes and families: Qwen2.5-3B-Base \cite{DBLP:journals/corr/abs-2412-15115}, Qwen2.5-7B-Base \cite{DBLP:journals/corr/abs-2412-15115}, and Llama3.1-8B-Base \cite{DBLP:journals/corr/abs-2407-21783}.

\paragraph{Baselines.}
To verify the effectiveness of BiRM, we consider a wide range of baselines, including the outcome reward model (ORM, \citealp{DBLP:journals/corr/abs-2110-14168}), process reward model (PRM, \citealp{DBLP:conf/iclr/LightmanKBEBLLS24}) and two variants of PRM: Math-Shepherd \cite{DBLP:conf/acl/WangLSXDLCWS24} and ER-PRM \cite{DBLP:journals/corr/abs-2412-11006}. Additionally, we include the results of greedy decoding and rule-based approaches, i.e. Majority Voting. We present more details in Appendix \ref{appendix:baseline}.

\paragraph{Implementation Details.}
In the SFT phase, we train our generators on the MATH subset of the MetaMath dataset \cite{DBLP:conf/iclr/YuJSYLZKLWL24} for two epochs, with a learning rate set to $1 \times 10^{-5}$. The global batch size is set to $256$. In the training phase, we use $225,000$ sampled solutions to train BiRM for one epoch based on the generator checkpoint with a learning rate of $5 \times 10^{-6}$. More details are provided in Appendix \ref{appendix:birm-train}.

\paragraph{Evaluation Metrics.}
We conduct a comprehensive evaluation of BiRM, considering both vanilla sampling and search strategies. Best-of-N (BoN) sampling is a commonly used evaluation metric for PRMs. It requires the model to score $N$ candidate solutions, with the highest-scoring solution selected as the final outcome. We also conduct beam search experiments to verify that BiRM can provide more promising guidance for LLM reasoning. In practice, BiRM follows Equation \ref{eq:BiRM-eval}, estimating both rewards and values to calculate final scores.
A detailed description is provided in Appendix \ref{appendix:eval}.

\subsection{Main Results}
\begin{table*}[t]
    \centering
    \resizebox{0.95\textwidth}{!}{
    \begin{tabular}{ll|ccccccccc}
    \toprule
    \multirow{2}{*}{\textbf{Models}} & \multirow{2}{*}{\textbf{\# Total Size}} & \multicolumn{3}{c|}{\textbf{GSM8K}} & \multicolumn{3}{c|}{\textbf{MATH-500}} & \multicolumn{3}{c}{\textbf{Gaokao2023}}  \\
    \cmidrule(lr){3-11}
    & & ORM & PRM & \multicolumn{1}{c|}{\blue{\textbf{BiRM}}} & ORM & PRM & \multicolumn{1}{c|}{\blue{\textbf{BiRM}}} & ORM & PRM & \multicolumn{1}{c}{\blue{\textbf{BiRM}}}  \\ \midrule
    
    \multirow{3}{*}{\textbf{Qwen2.5-3B}}
    & $K=4$ & $\mathbf{83.0}$ & $82.1$ & \multicolumn{1}{c|}{$82.8$} & $48.6$ & $49.3$ & \multicolumn{1}{c|}{$\mathbf{50.1}$} & $35.6$ & $34.9$ & \multicolumn{1}{c}{$\mathbf{36.1}$} \\
    & $K=8$ & $84.6$ & $83.9$ & \multicolumn{1}{c|}{$\mathbf{85.1}$} & $50.1$ & $50.9$ & \multicolumn{1}{c|}{$\mathbf{52.5}$} & $36.1$ & $\mathbf{37.9}$ & \multicolumn{1}{c}{$\mathbf{37.9}$} \\
    & $K=20$ & $86.7$ & $85.7$ & \multicolumn{1}{c|}{$\mathbf{86.9}$} & $53.0$ & $54.3$ & \multicolumn{1}{c|}{$\mathbf{55.0}$} & $37.7$ & $38.4$ & \multicolumn{1}{c}{$\mathbf{39.1}$} \\
    & $K=100$ & $87.5$ & $85.9$ & \multicolumn{1}{c|}{$\mathbf{87.6}$} & $53.0$ & $53.9$ & \multicolumn{1}{c|}{$\mathbf{55.1}$} & $38.1$ & $37.9$ & \multicolumn{1}{c}{$\mathbf{39.0}$} \\
    \midrule
    
    \multirow{3}{*}{\textbf{Qwen2.5-7B}}
    & $K=4$ & $86.2$ & $86.5$ & \multicolumn{1}{c|}{$\mathbf{87.0}$} & $55.7$ & $55.8$ & \multicolumn{1}{c|}{$\mathbf{57.1}$} & $42.8$ & $44.0$ & \multicolumn{1}{c}{$\mathbf{44.5}$} \\
    & $K=8$ & $88.6$ & $88.1$ & \multicolumn{1}{c|}{$\mathbf{89.4}$} & $58.3$ & $59.1$ & \multicolumn{1}{c|}{$\mathbf{60.1}$} & $44.2$ & $45.6$ & \multicolumn{1}{c}{$\mathbf{46.8}$} \\
    & $K=20$ & $90.4$ & $89.2$ & \multicolumn{1}{c|}{$\mathbf{90.6}$} & $59.1$ & $61.5$ & \multicolumn{1}{c|}{$\mathbf{62.3}$} & $45.5$ & $48.1$ & \multicolumn{1}{c}{$\mathbf{48.4}$} \\
    & $K=100$ & $91.2$ & $88.4$ & \multicolumn{1}{c|}{$\mathbf{91.7}$} & $60.1$ & $60.7$ & \multicolumn{1}{c|}{$\mathbf{62.5}$} & $46.8$ & $48.3$ & \multicolumn{1}{c}{$\mathbf{50.0}$} \\
    \midrule
    
    \multirow{3}{*}{\textbf{Llama3.1-8B}}
    & $K=4$ & $72.8$ & $71.7$ & \multicolumn{1}{c|}{$\mathbf{72.9}$} & $38.5$ & $39.9$ & \multicolumn{1}{c|}{$\mathbf{40.7}$} & $23.9$ & $25.1$ & \multicolumn{1}{c}{$\mathbf{25.4}$} \\
    & $K=8$ & $77.4$ & $75.9$ & \multicolumn{1}{c|}{$\mathbf{78.3}$} & $40.2$ & $40.1$ & \multicolumn{1}{c|}{$\mathbf{43.3}$} & $25.6$ & $26.6$ & \multicolumn{1}{c}{$\mathbf{27.5}$} \\
    & $K=20$ & $81.4$ & $79.2$ & \multicolumn{1}{c|}{$\mathbf{81.7}$} & $41.5$ & $42.1$ & \multicolumn{1}{c|}{$\mathbf{44.3}$} & $27.0$ & $28.6$ & \multicolumn{1}{c}{$\mathbf{29.2}$} \\
    & $K=100$ & $82.7$ & $80.3$ & \multicolumn{1}{c|}{$\mathbf{85.4}$} & $41.1$ & $42.3$ & \multicolumn{1}{c|}{$\mathbf{46.1}$} & $26.2$ & $29.6$ & \multicolumn{1}{c}{$\mathbf{30.7}$} \\
    \midrule
    
    \end{tabular}
    }
    \caption{Performance of beam search on GSM8K, MATH-500 and Gaokao2023 with three base models. ``\# Total Size'' stands for total sampling size $K$ in beam search and we report the best performance among all beam sizes. The results are reported as the average accuracy across three random seeds. The best results are marked in \textbf{bold}.}
    \label{tab:main-beam-search}
\end{table*}

\paragraph{BiRM exhibits more comprehensive and superior evaluations in BoN sampling.}
Table \ref{tab:main-result} presents a comparison of BoN accuracy across different supervision models on GSM8K, MATH-500, and the out-of-domain Gaokao2023 dataset. Our observations are as follows:
\textit{(1)} BiRM consistently outperforms vanilla ORM, PRM, and their variants on both GSM8K and MATH-500. For instance, BiRM trained on Llama3.1-8B outperforms PRM on GSM8K by $2.6\%$, while BiRM based on Qwen2.5-7B achieves an additional $5.0\%$ improvement on MATH-500.
\textit{(2)} BiRM exhibits better generalization ability. Since supervision models are trained solely on the query sets from GSM8K and MATH, Gaokao2023 serves as an out-of-domain (OOD) test set. BiRM-Qwen2.5-7B surpasses the finely labeled Math-Shepherd by $3.9\%$. In contrast, other supervision methods show fluctuating performance across different base models. 
\textit{(3)} As N increases, some supervision methods fail to provide consistent  supervision. For example, ORM trained on Qwen2.5-3B shows a decrease on Gaokao2023 from $37.2\%$ to $35.8\%$. In contrast, BiRM maintains a continuous increase in accuracy. We provide more detailed discussions in Section \ref{sec:analysis-positive}.

\paragraph{BiRM demonstrates more meaningful and promising guidance in search-based strategies.}
To fully demonstrate the superiority of BiRM’s bidirectional supervision capability, we conduct further experiments under search-based strategies . We run step-level beam search and choose vanilla ORM and PRM as baselines. The detailed algorithm is provided in Appendix \ref{appendix:beam-search}.
From Table \ref{tab:main-beam-search}, we can conclude that: \textit{(1)} BiRM achieves the highest accuracy in most cases. For example, on GSM8K, Qwen2.5-7B-BiRM achieves an accuracy of $89.4$ at  $K = 8$, which is a notable improvement over PRM’s $88.1\%$. \textit{(2)} As beam size increases, BiRM’s performance continues to improve. On the Llama3.1-8B base model, BiRM outperforms ORM by $2.8\%$ at  $K = 20$  and achieves a notable $5.0\%$ improvement at $K = 100$ in MATH-500 dataset. 
These results emphasize the valuable bidirectional supervision signals provided by BiRM, which significantly contributes to guiding the LLM toward more successful and promising final answers in solution searching. 

\section{Analysis and Discussions}

\subsection{Scaling Decline in BoN sampling}\label{sec:analysis-positive}
\begin{figure}[ht]
    \centering
    \includegraphics[width=0.48\textwidth]{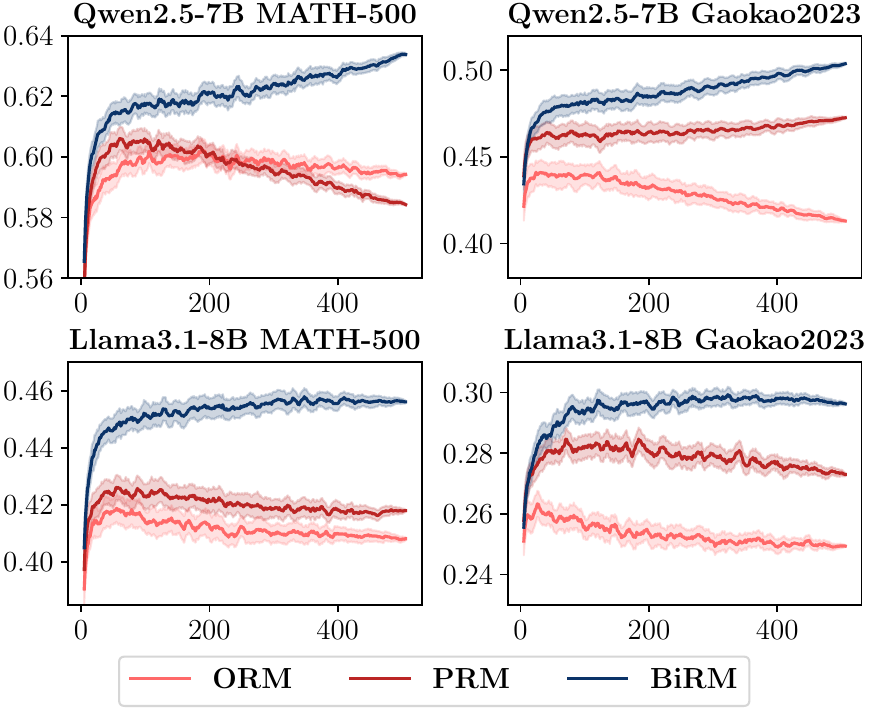}
    \caption{Scaling decline phenomenon in Best-of-N sampling. We present the BoN accuracy results across five random seeds. For better visualization, we apply the moving average with a window size of $10$. }
    \vspace{-10pt}
    \label{fig:analysis-scaling-decline}
\end{figure}

We conduct a further analysis of the scaling decline phenomenon in our main results. The complete BoN accuracy curve, shown in Figure \ref{fig:analysis-scaling-decline}, is plotted for $N$ ranging from $1$ to $512$.  As $N$ increases, we observe that BiRM shows a consistent improvement. In contrast, the post-verification accuracy of vanilla ORM and PRM plateaus and even declines, which contradicts our intuition learned from the test-time scaling laws \cite{DBLP:journals/corr/abs-2408-03314}.

\begin{table}[ht]
    \centering
    \resizebox{\linewidth}{!}{
        \begin{tabular}{ll | cccc}
        \toprule
        \multirow{2}{*}{\textbf{Models}} & \multirow{2}{*}{\textbf{Methods}} & \multicolumn{2}{c|}{\textbf{MATH-500}} & \multicolumn{2}{c}{\textbf{Gaokao2023}} \\
        \cmidrule(lr){3-6}
    
        & & @\textit{128} & \multicolumn{1}{c|}{@\textit{512}} & @\textit{128} & @\textit{512} \\ 
    
        \midrule
        \multirow{3}{*}{\textbf{Qwen2.5-7B}} & + Outcome & $61.8$ & \multicolumn{1}{c|}{$61.1$} & $46.8$ & $49.4$ \\
        & + MS. (Hard) & $62.1$ & \multicolumn{1}{c|}{$62.8$} & $47.3$ & $49.7$ \\
        & + MS. (Soft) & $\mathbf{62.4}$ & \multicolumn{1}{c|}{$\mathbf{63.4}$} & $\mathbf{47.7}$ & $\mathbf{50.4}$\\
        \midrule
        \multirow{3}{*}{\textbf{Llama3.1-8B}} & + Outcome & $44.9$ & \multicolumn{1}{c|}{$44.2$} & $29.0$ & $\mathbf{29.6}$ \\
        & + MS. (Hard) & $45.1$ & \multicolumn{1}{c|}{$45.4$} & $29.2$ & $29.4$ \\
        & + MS. (Soft) & $\mathbf{45.4}$ & \multicolumn{1}{c|}{$\mathbf{45.6}$} & $\mathbf{29.4}$ & $\mathbf{29.6}$ \\

        \bottomrule
        \end{tabular}
    }
    \caption{Different value label annotation strategies for BiRM. ``Outcome'' stands for Outcome-supervised estimation. ``MS. (Hard)'' and ``MS. (Soft)'' represents Math-Shepherd hard and soft estimation respectively.}
    \vspace{-5pt}
    \label{tab:value-label-annotation}
\end{table}

We attribute this decline to verifier failures. Imperfect verifiers misrank candidates, erroneously classifying positive samples as negative. As the sample size increases, this misjudgment becomes more pronounced. Traditional PRMs exhibit a one-directional scoring nature, limiting their ability to evaluate candidates from a comprehensive perspective. In contrast, BiRM estimates both rewards and values, providing more reliable supervision signals.

\subsection{Annotation Strategies for Value Labels}

As discussed in Section \ref{sec:value-label}, we explore various strategies for annotating precise value labels. We aim to demonstrate that our method has good orthogonality with existing annotation strategies.

Table \ref{tab:value-label-annotation} presents the accuracy of BiRM in BoN sampling under different strategies. We can conclude that: \textit{(1)} More accurate annotations lead to greater improvements. The mash-shepherd soft estimation, which uses the potential success probability of intermediate steps as explicit labels, offers the finest granularity and achieves the best performance. In contrast, outcome-supervised estimation, which relies on outcome labels for implicit learning, exhibits greater variability. \textit{(2)} Even the weakest method, outcome-supervised estimation, shows a notable improvement over PRM. This highlights the flexibility and applicability of BiRM.

\subsection{Orthogonality to Existing PRMs}
\begin{figure}[t]
    \centering
    \includegraphics[width=0.96\linewidth]{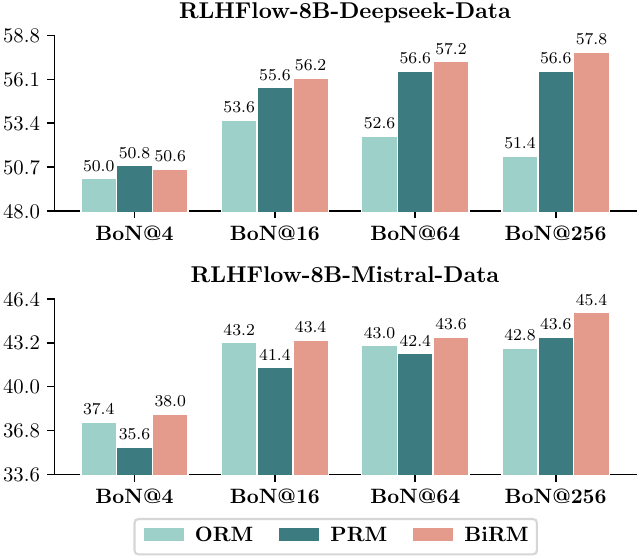}
    \caption{Performance comparison of ORM, PRM and BiRM under BoN sampling. The base models are open-source RLHFlow-8B-Deepseek-Data and RLHFlow-8B-Mistral-Data \cite{xiong2024rlhflowmath}. We follow Equation~\ref{eq:BiRM-eval} to calculate the BiRM score at test-time.
    }
    \label{fig:openPRM}
\end{figure}
To further demonstrate the generalization ability of our method, we conduct experiments using several existing open-source reward models. We select ORMs and PRMs trained by RLHFlow \cite{xiong2024rlhflowmath, rlhflow} as baselines and reuse the $N$ sampled solutions they provided for testing. Then we follow Equation \ref{eq:BiRM-eval} to calculate the BiRM scores for BoN sampling.

Experiment results in Figure \ref{fig:openPRM} clearly reveal that BiRM consistently outperforms both ORM and PRM across different values of $N$, maintaining a consistent upward trend. Furthermore, this trend expands at larger sampling sizes, where BiRM maintains its lead, reaching an accuracy of $57.8\%$ at BoN@256, compared to PRM's $56.6\%$ and ORM's $51.4\%$, respectively. These findings indicate the reliability and generalization of BiRM when using existing open-source reward models.

\subsection{Query Scaling or Response Scaling}
\begin{table}[ht]
    \centering
    \resizebox{\linewidth}{!}{
        \begin{tabular}{cc | cccc}
        \toprule
        \multirow{2}{*}{\textbf{\# Query}} & \multirow{2}{*}{\textbf{\# Resp.}} & \multicolumn{2}{c|}{\textbf{MATH-500}} & \multicolumn{2}{c}{\textbf{Gaokao2023}} \\
        \cmidrule(lr){3-6}
    
        & & @\textit{128} & \multicolumn{1}{c|}{@\textit{512}} & @\textit{128} & @\textit{512} \\ 
    
        \midrule
        \multirow{3}{*}{$15,000$} & $\times 30$  & $61.3$ & \multicolumn{1}{c|}{$61.6$} & $47.3$ & $48.3$ \\
               & $\times 15$  & $\mathbf{62.0}$ & \multicolumn{1}{c|}{$\mathbf{63.0}$} & $\mathbf{46.8}$ & $\mathbf{49.4}$ \\
               & $\times 8$  & $61.3$ & \multicolumn{1}{c|}{$61.2$} & $46.4$ & $46.8$ \\
        \midrule
        $7,500$ & $\times 15$  & $59.0$ & \multicolumn{1}{c|}{$58.8$} & $45.4$ & $44.7$ \\
        \midrule
        $3,750$ &  $\times 30$ & $57.9$ & \multicolumn{1}{c|}{$58.2$} & $43.4$ & $42.8$ \\
        \bottomrule
        \end{tabular}
    }
    \caption{Training data scaling for queries and responses. The base model is Qwen2.5-7B and we use outcome-supervised estimation for simplicity.}
    \vspace{-8pt}
    \label{tab:queryxresponse-scaling}
\end{table}
We also explore a key issue in training supervision models: which matters more, query scaling or response scaling? 

We first fix the number of queries and use the original GSM8K and MATH datasets, which contain approximately $15,000$ queries. We then test BiRM’s performance with response sizes of $8$,$15$, and $30$. The results in Table \ref{tab:queryxresponse-scaling} reveal that BiRM performs best when the response size is $15$ on both datasets. 
The possible reason is that when the number of responses is too low, BiRM cannot learn sufficient and diverse supervision signals. On the other hand, the model struggles with overly similar data patterns per query when $\text{\# Resp.} = 30$, leading to overfitting.

Furthermore, we control the total size of the training dataset. Specifically, we conduct experiments with three following settings: $15,000  \times 8$, $7,500  \times 15$, and $3,750  \times 30$. The results demonstrate that BiRM performs best with the $15,000 \times 8$ configuration. Additionally, we observe that models with fewer queries go through more severe degradation when facing OOD test sets. In the MATH-500 experiments, the gap between the $7,500 \times 15$ and $ 3,750 \times 30$ settings ranges from 0.6\% to 1.1\%, but this gap significantly widens to 2.0\% on the Gaokao2023 benchmark. To sum up, we believe that maintaining an appropriate response size while scaling the number of queries is critical to training process supervision models. We hope this provides valuable insights to the community.

\section{Conclusion}
In this work, we introduce BiRM, a novel process supervision model for large language models (LLMs), inspired by the A* algorithm. BiRM provides bidirectional supervision signals, evaluating both the correctness of reasoning steps taken so far and the probability of reaching correct answers in the future. Our extensive experiments demonstrate the effectiveness of BiRM across various mathematical reasoning tasks, outperforming existing supervision models like ORM and PRM. Through detailed analysis, we highlight the strengths of BiRM in guiding the search process and improving solution re-ranking. We hope that our approach contributes valuable insights to the field of process supervision and opens avenues for future research in enhancing LLM-based reasoning.

\section*{Limitations}
Our work has some limitations, which we leave for future work to address: \textit{(1)} High computational cost in test-time searching. In order to improve the performance of LLMs at test-time, we employ vanilla sampling and search-based strategies for solution searching. However, this process requires a significant amount of computational resources. In our work, we use vLLM \cite{DBLP:conf/sosp/KwonLZ0ZY0ZS23} to alleviate this limitation. Besides, we also observe that search-based strategies sometimes perform worse than repeated sampling due to verifier failures \cite{yu2025scaling}, even under the same computational budget. We will explore this problem in the future. \textit{(2)} Generalization across different data patterns and base models. In our experiments, we train our generators and supervision models based on the same base models, ensuring the same data patterns. However, in practical scenarios, an optimal supervision model should be independent of the data pattern and capable of supervising different kinds of reasoning paths. We hope our work provides insights to the community and contributes to the development of more robust and generalized supervision models.

\bibliography{custom}

\begin{thebibliography}{37}
\providecommand{\natexlab}[1]{#1}

\bibitem[{Ankner et~al.(2024)Ankner, Paul, Cui, Chang, and Ammanabrolu}]{DBLP:journals/corr/abs-2408-11791}
Zachary Ankner, Mansheej Paul, Brandon Cui, Jonathan~D. Chang, and Prithviraj Ammanabrolu. 2024.
\newblock \href {https://doi.org/10.48550/ARXIV.2408.11791} {Critique-out-loud reward models}.
\newblock \emph{CoRR}, abs/2408.11791.

\bibitem[{Brown et~al.(2024)Brown, Juravsky, Ehrlich, Clark, Le, R{\'{e}}, and Mirhoseini}]{DBLP:journals/corr/abs-2407-21787}
Bradley C.~A. Brown, Jordan Juravsky, Ryan~Saul Ehrlich, Ronald Clark, Quoc~V. Le, Christopher R{\'{e}}, and Azalia Mirhoseini. 2024.
\newblock \href {https://doi.org/10.48550/ARXIV.2407.21787} {Large language monkeys: Scaling inference compute with repeated sampling}.
\newblock \emph{CoRR}, abs/2407.21787.

\bibitem[{Cobbe et~al.(2021)Cobbe, Kosaraju, Bavarian, Chen, Jun, Kaiser, Plappert, Tworek, Hilton, Nakano, Hesse, and Schulman}]{DBLP:journals/corr/abs-2110-14168}
Karl Cobbe, Vineet Kosaraju, Mohammad Bavarian, Mark Chen, Heewoo Jun, Lukasz Kaiser, Matthias Plappert, Jerry Tworek, Jacob Hilton, Reiichiro Nakano, Christopher Hesse, and John Schulman. 2021.
\newblock \href {https://arxiv.org/abs/2110.14168} {Training verifiers to solve math word problems}.
\newblock \emph{CoRR}, abs/2110.14168.

\bibitem[{DeepSeek-AI et~al.(2025)DeepSeek-AI, Guo, Yang, Zhang, Song, Zhang, Xu, Zhu, Ma, Wang, Bi, Zhang, Yu, Wu, Wu, Gou, Shao, Li, Gao, Liu, Xue, Wang, Wu, Feng, Lu, Zhao, Deng, Zhang, Ruan, Dai, Chen, Ji, Li, Lin, Dai, Luo, Hao, Chen, Li, Zhang, Bao, Xu, Wang, Ding, Xin, Gao, Qu, Li, Guo, Li, Wang, Chen, Yuan, Qiu, Li, Cai, Ni, Liang, Chen, Dong, Hu, Gao, Guan, Huang, Yu, Wang, Zhang, Zhao, Wang, Zhang, Xu, Xia, Zhang, Zhang, Tang, Li, Wang, Li, Tian, Huang, Zhang, Wang, Chen, Du, Ge, Zhang, Pan, Wang, Chen, Jin, Chen, Lu, Zhou, Chen, Ye, Wang, Yu, Zhou, Pan, Li, Zhou, Wu, Ye, Yun, Pei, Sun, Wang, Zeng, Zhao, Liu, Liang, Gao, Yu, Zhang, Xiao, An, Liu, Wang, Chen, Nie, Cheng, Liu, Xie, Liu, Yang, Li, Su, Lin, Li, Jin, Shen, Chen, Sun, Wang, Song, Zhou, Wang, Shan, Li, Wang, Wei, Zhang, Xu, Li, Zhao, Sun, Wang, Yu, Zhang, Shi, Xiong, He, Piao, Wang, Tan, Ma, Liu, Guo, Ou, Wang, Gong, Zou, He, Xiong, Luo, You, Liu, Zhou, Zhu, Xu, Huang, Li, Zheng, Zhu, Ma, Tang, Zha, Yan, Ren, Ren, Sha, Fu, Xu, Xie, Zhang,
  Hao, Ma, Yan, Wu, Gu, Zhu, Liu, Li, Xie, Song, Pan, Huang, Xu, Zhang, and Zhang}]{deepseekai2025deepseekr1incentivizingreasoningcapability}
DeepSeek-AI, Daya Guo, Dejian Yang, Haowei Zhang, Junxiao Song, Ruoyu Zhang, Runxin Xu, Qihao Zhu, Shirong Ma, Peiyi Wang, Xiao Bi, Xiaokang Zhang, Xingkai Yu, Yu~Wu, Z.~F. Wu, Zhibin Gou, Zhihong Shao, Zhuoshu Li, Ziyi Gao, Aixin Liu, Bing Xue, Bingxuan Wang, Bochao Wu, Bei Feng, Chengda Lu, Chenggang Zhao, Chengqi Deng, Chenyu Zhang, Chong Ruan, Damai Dai, Deli Chen, Dongjie Ji, Erhang Li, Fangyun Lin, Fucong Dai, Fuli Luo, Guangbo Hao, Guanting Chen, Guowei Li, H.~Zhang, Han Bao, Hanwei Xu, Haocheng Wang, Honghui Ding, Huajian Xin, Huazuo Gao, Hui Qu, Hui Li, Jianzhong Guo, Jiashi Li, Jiawei Wang, Jingchang Chen, Jingyang Yuan, Junjie Qiu, Junlong Li, J.~L. Cai, Jiaqi Ni, Jian Liang, Jin Chen, Kai Dong, Kai Hu, Kaige Gao, Kang Guan, Kexin Huang, Kuai Yu, Lean Wang, Lecong Zhang, Liang Zhao, Litong Wang, Liyue Zhang, Lei Xu, Leyi Xia, Mingchuan Zhang, Minghua Zhang, Minghui Tang, Meng Li, Miaojun Wang, Mingming Li, Ning Tian, Panpan Huang, Peng Zhang, Qiancheng Wang, Qinyu Chen, Qiushi Du, Ruiqi Ge, Ruisong
  Zhang, Ruizhe Pan, Runji Wang, R.~J. Chen, R.~L. Jin, Ruyi Chen, Shanghao Lu, Shangyan Zhou, Shanhuang Chen, Shengfeng Ye, Shiyu Wang, Shuiping Yu, Shunfeng Zhou, Shuting Pan, S.~S. Li, Shuang Zhou, Shaoqing Wu, Shengfeng Ye, Tao Yun, Tian Pei, Tianyu Sun, T.~Wang, Wangding Zeng, Wanjia Zhao, Wen Liu, Wenfeng Liang, Wenjun Gao, Wenqin Yu, Wentao Zhang, W.~L. Xiao, Wei An, Xiaodong Liu, Xiaohan Wang, Xiaokang Chen, Xiaotao Nie, Xin Cheng, Xin Liu, Xin Xie, Xingchao Liu, Xinyu Yang, Xinyuan Li, Xuecheng Su, Xuheng Lin, X.~Q. Li, Xiangyue Jin, Xiaojin Shen, Xiaosha Chen, Xiaowen Sun, Xiaoxiang Wang, Xinnan Song, Xinyi Zhou, Xianzu Wang, Xinxia Shan, Y.~K. Li, Y.~Q. Wang, Y.~X. Wei, Yang Zhang, Yanhong Xu, Yao Li, Yao Zhao, Yaofeng Sun, Yaohui Wang, Yi~Yu, Yichao Zhang, Yifan Shi, Yiliang Xiong, Ying He, Yishi Piao, Yisong Wang, Yixuan Tan, Yiyang Ma, Yiyuan Liu, Yongqiang Guo, Yuan Ou, Yuduan Wang, Yue Gong, Yuheng Zou, Yujia He, Yunfan Xiong, Yuxiang Luo, Yuxiang You, Yuxuan Liu, Yuyang Zhou, Y.~X. Zhu,
  Yanhong Xu, Yanping Huang, Yaohui Li, Yi~Zheng, Yuchen Zhu, Yunxian Ma, Ying Tang, Yukun Zha, Yuting Yan, Z.~Z. Ren, Zehui Ren, Zhangli Sha, Zhe Fu, Zhean Xu, Zhenda Xie, Zhengyan Zhang, Zhewen Hao, Zhicheng Ma, Zhigang Yan, Zhiyu Wu, Zihui Gu, Zijia Zhu, Zijun Liu, Zilin Li, Ziwei Xie, Ziyang Song, Zizheng Pan, Zhen Huang, Zhipeng Xu, Zhongyu Zhang, and Zhen Zhang. 2025.
\newblock \href {https://arxiv.org/abs/2501.12948} {Deepseek-r1: Incentivizing reasoning capability in llms via reinforcement learning}.
\newblock \emph{Preprint}, arXiv:2501.12948.

\bibitem[{DeepSeek{-}AI et~al.(2024)DeepSeek{-}AI, Liu, Feng, Xue, Wang, Wu, Lu, Zhao, Deng, Zhang, Ruan, Dai, Guo, Yang, Chen, Ji, Li, Lin, Dai, Luo, Hao, Chen, Li, Zhang, Bao, Xu, Wang, Zhang, Ding, Xin, Gao, Li, Qu, Cai, Liang, Guo, Ni, Li, Wang, Chen, Chen, Yuan, Qiu, Li, Song, Dong, Hu, Gao, Guan, Huang, Yu, Wang, Zhang, Xu, Xia, Zhao, Wang, Zhang, Li, Wang, Zhang, Zhang, Tang, Li, Tian, Huang, Wang, Zhang, Wang, Zhu, Chen, Du, Chen, Jin, Ge, Zhang, Pan, Wang, Xu, Zhang, Chen, Li, Lu, Zhou, Chen, Wu, Ye, Ye, Ma, Wang, Zhou, Yu, Zhou, Pan, Wang, Yun, Pei, Sun, Xiao, and Zeng}]{DBLP:journals/corr/abs-2412-19437}
DeepSeek{-}AI, Aixin Liu, Bei Feng, Bing Xue, Bingxuan Wang, Bochao Wu, Chengda Lu, Chenggang Zhao, Chengqi Deng, Chenyu Zhang, Chong Ruan, Damai Dai, Daya Guo, Dejian Yang, Deli Chen, Dongjie Ji, Erhang Li, Fangyun Lin, Fucong Dai, Fuli Luo, Guangbo Hao, Guanting Chen, Guowei Li, H.~Zhang, Han Bao, Hanwei Xu, Haocheng Wang, Haowei Zhang, Honghui Ding, Huajian Xin, Huazuo Gao, Hui Li, Hui Qu, J.~L. Cai, Jian Liang, Jianzhong Guo, Jiaqi Ni, Jiashi Li, Jiawei Wang, Jin Chen, Jingchang Chen, Jingyang Yuan, Junjie Qiu, Junlong Li, Junxiao Song, Kai Dong, Kai Hu, Kaige Gao, Kang Guan, Kexin Huang, Kuai Yu, Lean Wang, Lecong Zhang, Lei Xu, Leyi Xia, Liang Zhao, Litong Wang, Liyue Zhang, Meng Li, Miaojun Wang, Mingchuan Zhang, Minghua Zhang, Minghui Tang, Mingming Li, Ning Tian, Panpan Huang, Peiyi Wang, Peng Zhang, Qiancheng Wang, Qihao Zhu, Qinyu Chen, Qiushi Du, R.~J. Chen, R.~L. Jin, Ruiqi Ge, Ruisong Zhang, Ruizhe Pan, Runji Wang, Runxin Xu, Ruoyu Zhang, Ruyi Chen, S.~S. Li, Shanghao Lu, Shangyan Zhou,
  Shanhuang Chen, Shaoqing Wu, Shengfeng Ye, Shengfeng Ye, Shirong Ma, Shiyu Wang, Shuang Zhou, Shuiping Yu, Shunfeng Zhou, Shuting Pan, T.~Wang, Tao Yun, Tian Pei, Tianyu Sun, W.~L. Xiao, and Wangding Zeng. 2024.
\newblock \href {https://doi.org/10.48550/ARXIV.2412.19437} {Deepseek-v3 technical report}.
\newblock \emph{CoRR}, abs/2412.19437.

\bibitem[{Dong et~al.(2024)Dong, Xiong, Pang, Wang, Zhao, Zhou, Jiang, Sahoo, Xiong, and Zhang}]{rlhflow}
Hanze Dong, Wei Xiong, Bo~Pang, Haoxiang Wang, Han Zhao, Yingbo Zhou, Nan Jiang, Doyen Sahoo, Caiming Xiong, and Tong Zhang. 2024.
\newblock \href {https://arxiv.org/abs/2405.07863} {Rlhf workflow: From reward modeling to online rlhf}.
\newblock \emph{Preprint}, arXiv:2405.07863.

\bibitem[{Dubey et~al.(2024)Dubey, Jauhri, Pandey, Kadian, Al{-}Dahle, Letman, Mathur, Schelten, Yang, Fan, Goyal, Hartshorn, Yang, Mitra, Sravankumar, Korenev, Hinsvark, Rao, Zhang, Rodriguez, Gregerson, Spataru, Rozi{\`{e}}re, Biron, Tang, Chern, Caucheteux, Nayak, Bi, Marra, McConnell, Keller, Touret, Wu, Wong, Ferrer, Nikolaidis, Allonsius, Song, Pintz, Livshits, Esiobu, Choudhary, Mahajan, Garcia{-}Olano, Perino, Hupkes, Lakomkin, AlBadawy, Lobanova, Dinan, Smith, Radenovic, Zhang, Synnaeve, Lee, Anderson, Nail, Mialon, Pang, Cucurell, Nguyen, Korevaar, Xu, Touvron, Zarov, Ibarra, Kloumann, Misra, Evtimov, Copet, Lee, Geffert, Vranes, Park, Mahadeokar, Shah, van~der Linde, Billock, Hong, Lee, Fu, Chi, Huang, Liu, Wang, Yu, Bitton, Spisak, Park, Rocca, Johnstun, Saxe, Jia, Alwala, Upasani, Plawiak, Li, Heafield, Stone, and et~al.}]{DBLP:journals/corr/abs-2407-21783}
Abhimanyu Dubey, Abhinav Jauhri, Abhinav Pandey, Abhishek Kadian, Ahmad Al{-}Dahle, Aiesha Letman, Akhil Mathur, Alan Schelten, Amy Yang, Angela Fan, Anirudh Goyal, Anthony Hartshorn, Aobo Yang, Archi Mitra, Archie Sravankumar, Artem Korenev, Arthur Hinsvark, Arun Rao, Aston Zhang, Aur{\'{e}}lien Rodriguez, Austen Gregerson, Ava Spataru, Baptiste Rozi{\`{e}}re, Bethany Biron, Binh Tang, Bobbie Chern, Charlotte Caucheteux, Chaya Nayak, Chloe Bi, Chris Marra, Chris McConnell, Christian Keller, Christophe Touret, Chunyang Wu, Corinne Wong, Cristian~Canton Ferrer, Cyrus Nikolaidis, Damien Allonsius, Daniel Song, Danielle Pintz, Danny Livshits, David Esiobu, Dhruv Choudhary, Dhruv Mahajan, Diego Garcia{-}Olano, Diego Perino, Dieuwke Hupkes, Egor Lakomkin, Ehab AlBadawy, Elina Lobanova, Emily Dinan, Eric~Michael Smith, Filip Radenovic, Frank Zhang, Gabriel Synnaeve, Gabrielle Lee, Georgia~Lewis Anderson, Graeme Nail, Gr{\'{e}}goire Mialon, Guan Pang, Guillem Cucurell, Hailey Nguyen, Hannah Korevaar, Hu~Xu, Hugo
  Touvron, Iliyan Zarov, Imanol~Arrieta Ibarra, Isabel~M. Kloumann, Ishan Misra, Ivan Evtimov, Jade Copet, Jaewon Lee, Jan Geffert, Jana Vranes, Jason Park, Jay Mahadeokar, Jeet Shah, Jelmer van~der Linde, Jennifer Billock, Jenny Hong, Jenya Lee, Jeremy Fu, Jianfeng Chi, Jianyu Huang, Jiawen Liu, Jie Wang, Jiecao Yu, Joanna Bitton, Joe Spisak, Jongsoo Park, Joseph Rocca, Joshua Johnstun, Joshua Saxe, Junteng Jia, Kalyan~Vasuden Alwala, Kartikeya Upasani, Kate Plawiak, Ke~Li, Kenneth Heafield, Kevin Stone, and et~al. 2024.
\newblock \href {https://doi.org/10.48550/ARXIV.2407.21783} {The llama 3 herd of models}.
\newblock \emph{CoRR}, abs/2407.21783.

\bibitem[{Hart et~al.(1968)Hart, Nilsson, and Raphael}]{DBLP:journals/tssc/HartNR68}
Peter~E. Hart, Nils~J. Nilsson, and Bertram Raphael. 1968.
\newblock \href {https://doi.org/10.1109/TSSC.1968.300136} {A formal basis for the heuristic determination of minimum cost paths}.
\newblock \emph{{IEEE} Trans. Syst. Sci. Cybern.}, 4(2):100--107.

\bibitem[{Kwon et~al.(2023)Kwon, Li, Zhuang, Sheng, Zheng, Yu, Gonzalez, Zhang, and Stoica}]{DBLP:conf/sosp/KwonLZ0ZY0ZS23}
Woosuk Kwon, Zhuohan Li, Siyuan Zhuang, Ying Sheng, Lianmin Zheng, Cody~Hao Yu, Joseph Gonzalez, Hao Zhang, and Ion Stoica. 2023.
\newblock \href {https://doi.org/10.1145/3600006.3613165} {Efficient memory management for large language model serving with pagedattention}.
\newblock In \emph{Proceedings of the 29th Symposium on Operating Systems Principles, {SOSP} 2023, Koblenz, Germany, October 23-26, 2023}, pages 611--626. {ACM}.

\bibitem[{Liao et~al.(2024)Liao, Li, Luo, Wu, and Fan}]{DBLP:conf/acl/Liao0LW024}
Minpeng Liao, Chengxi Li, Wei Luo, Jing Wu, and Kai Fan. 2024.
\newblock \href {https://doi.org/10.18653/V1/2024.FINDINGS-ACL.53} {{MARIO:} math reasoning with code interpreter output - {A} reproducible pipeline}.
\newblock In \emph{Findings of the Association for Computational Linguistics, {ACL} 2024, Bangkok, Thailand and virtual meeting, August 11-16, 2024}, pages 905--924. Association for Computational Linguistics.

\bibitem[{Lightman et~al.(2024)Lightman, Kosaraju, Burda, Edwards, Baker, Lee, Leike, Schulman, Sutskever, and Cobbe}]{DBLP:conf/iclr/LightmanKBEBLLS24}
Hunter Lightman, Vineet Kosaraju, Yuri Burda, Harrison Edwards, Bowen Baker, Teddy Lee, Jan Leike, John Schulman, Ilya Sutskever, and Karl Cobbe. 2024.
\newblock \href {https://openreview.net/forum?id=v8L0pN6EOi} {Let's verify step by step}.
\newblock In \emph{The Twelfth International Conference on Learning Representations, {ICLR} 2024, Vienna, Austria, May 7-11, 2024}. OpenReview.net.

\bibitem[{Liu et~al.(2025)Liu, Yao, Min, Cao, Hou, and Li}]{liu2025pairwise}
Yantao Liu, Zijun Yao, Rui Min, Yixin Cao, Lei Hou, and Juanzi Li. 2025.
\newblock Pairwise rm: Perform best-of-n sampling with knockout tournament.
\newblock \emph{arXiv preprint arXiv:2501.13007}.

\bibitem[{Luo et~al.(2024)Luo, Liu, Liu, Phatale, Lara, Li, Shu, Zhu, Meng, Sun, and Rastogi}]{DBLP:journals/corr/abs-2406-06592}
Liangchen Luo, Yinxiao Liu, Rosanne Liu, Samrat Phatale, Harsh Lara, Yunxuan Li, Lei Shu, Yun Zhu, Lei Meng, Jiao Sun, and Abhinav Rastogi. 2024.
\newblock \href {https://doi.org/10.48550/ARXIV.2406.06592} {Improve mathematical reasoning in language models by automated process supervision}.
\newblock \emph{CoRR}, abs/2406.06592.

\bibitem[{OpenAI(2024{\natexlab{a}})}]{gpt4o}
OpenAI. 2024{\natexlab{a}}.
\newblock \href {https://openai.com/index/hello-gpt-4o/} {{GPT}-4o}.

\bibitem[{OpenAI(2024{\natexlab{b}})}]{openai-o1}
OpenAI. 2024{\natexlab{b}}.
\newblock \href {https://openai.com/o1/} {Introducing openai o1}.

\bibitem[{Snell et~al.(2024)Snell, Lee, Xu, and Kumar}]{DBLP:journals/corr/abs-2408-03314}
Charlie Snell, Jaehoon Lee, Kelvin Xu, and Aviral Kumar. 2024.
\newblock \href {https://doi.org/10.48550/ARXIV.2408.03314} {Scaling {LLM} test-time compute optimally can be more effective than scaling model parameters}.
\newblock \emph{CoRR}, abs/2408.03314.

\bibitem[{Song et~al.(2025)Song, Su, Qu, Zhou, and Cheng}]{song2025prmbench}
Mingyang Song, Zhaochen Su, Xiaoye Qu, Jiawei Zhou, and Yu~Cheng. 2025.
\newblock Prmbench: A fine-grained and challenging benchmark for process-level reward models.
\newblock \emph{arXiv preprint arXiv:2501.03124}.

\bibitem[{Stroebl et~al.(2024)Stroebl, Kapoor, and Narayanan}]{stroebl2024inference}
Benedikt Stroebl, Sayash Kapoor, and Arvind Narayanan. 2024.
\newblock Inference scaling laws: The limits of llm resampling with imperfect verifiers.
\newblock \emph{arXiv preprint arXiv:2411.17501}.

\bibitem[{Tong et~al.(2024)Tong, Zhang, Wang, Wu, and He}]{DBLP:conf/nips/TongZWWH24}
Yuxuan Tong, Xiwen Zhang, Rui Wang, Ruidong Wu, and Junxian He. 2024.
\newblock \href {http://papers.nips.cc/paper\_files/paper/2024/hash/0ef1afa0daa888d695dcd5e9513bafa3-Abstract-Conference.html} {Dart-math: Difficulty-aware rejection tuning for mathematical problem-solving}.
\newblock In \emph{Advances in Neural Information Processing Systems 38: Annual Conference on Neural Information Processing Systems 2024, NeurIPS 2024, Vancouver, BC, Canada, December 10 - 15, 2024}.

\bibitem[{Uesato et~al.(2022)Uesato, Kushman, Kumar, Song, Siegel, Wang, Creswell, Irving, and Higgins}]{DBLP:journals/corr/abs-2211-14275}
Jonathan Uesato, Nate Kushman, Ramana Kumar, H.~Francis Song, Noah~Y. Siegel, Lisa Wang, Antonia Creswell, Geoffrey Irving, and Irina Higgins. 2022.
\newblock \href {https://doi.org/10.48550/ARXIV.2211.14275} {Solving math word problems with process- and outcome-based feedback}.
\newblock \emph{CoRR}, abs/2211.14275.

\bibitem[{Wan et~al.(2024)Wan, Feng, Wen, McAleer, Wen, Zhang, and Wang}]{DBLP:conf/icml/WanFWM00024}
Ziyu Wan, Xidong Feng, Muning Wen, Stephen~Marcus McAleer, Ying Wen, Weinan Zhang, and Jun Wang. 2024.
\newblock \href {https://openreview.net/forum?id=C4OpREezgj} {Alphazero-like tree-search can guide large language model decoding and training}.
\newblock In \emph{Forty-first International Conference on Machine Learning, {ICML} 2024, Vienna, Austria, July 21-27, 2024}. OpenReview.net.

\bibitem[{Wang et~al.(2024{\natexlab{a}})Wang, Deng, Lv, Liang, He, Yan, and An}]{DBLP:journals/corr/abs-2406-14283}
Chaojie Wang, Yanchen Deng, Zhiyi Lv, Zeng Liang, Jujie He, Shuicheng Yan, and Bo~An. 2024{\natexlab{a}}.
\newblock \href {https://doi.org/10.48550/ARXIV.2406.14283} {Q*: Improving multi-step reasoning for llms with deliberative planning}.
\newblock \emph{CoRR}, abs/2406.14283.

\bibitem[{Wang et~al.(2024{\natexlab{b}})Wang, Li, Shao, Xu, Dai, Li, Chen, Wu, and Sui}]{DBLP:conf/acl/WangLSXDLCWS24}
Peiyi Wang, Lei Li, Zhihong Shao, Runxin Xu, Damai Dai, Yifei Li, Deli Chen, Yu~Wu, and Zhifang Sui. 2024{\natexlab{b}}.
\newblock \href {https://doi.org/10.18653/V1/2024.ACL-LONG.510} {Math-shepherd: Verify and reinforce llms step-by-step without human annotations}.
\newblock In \emph{Proceedings of the 62nd Annual Meeting of the Association for Computational Linguistics (Volume 1: Long Papers), {ACL} 2024, Bangkok, Thailand, August 11-16, 2024}, pages 9426--9439. Association for Computational Linguistics.

\bibitem[{Wang et~al.(2025)Wang, Yang, Wang, and Wei}]{wang2025examining}
Yu~Wang, Nan Yang, Liang Wang, and Furu Wei. 2025.
\newblock Examining false positives under inference scaling for mathematical reasoning.
\newblock \emph{arXiv preprint arXiv:2502.06217}.

\bibitem[{Wei et~al.(2022)Wei, Wang, Schuurmans, Bosma, Ichter, Xia, Chi, Le, and Zhou}]{DBLP:conf/nips/Wei0SBIXCLZ22}
Jason Wei, Xuezhi Wang, Dale Schuurmans, Maarten Bosma, Brian Ichter, Fei Xia, Ed~H. Chi, Quoc~V. Le, and Denny Zhou. 2022.
\newblock \href {http://papers.nips.cc/paper\_files/paper/2022/hash/9d5609613524ecf4f15af0f7b31abca4-Abstract-Conference.html} {Chain-of-thought prompting elicits reasoning in large language models}.
\newblock In \emph{Advances in Neural Information Processing Systems 35: Annual Conference on Neural Information Processing Systems 2022, NeurIPS 2022, New Orleans, LA, USA, November 28 - December 9, 2022}.

\bibitem[{Wu et~al.(2024)Wu, Sun, Li, Welleck, and Yang}]{wu2024inference}
Yangzhen Wu, Zhiqing Sun, Shanda Li, Sean Welleck, and Yiming Yang. 2024.
\newblock Inference scaling laws: An empirical analysis of compute-optimal inference for llm problem-solving.
\newblock In \emph{The 4th Workshop on Mathematical Reasoning and AI at NeurIPS'24}.

\bibitem[{Xiong et~al.(2024)Xiong, Zhang, Jiang, and Zhang}]{xiong2024rlhflowmath}
Wei Xiong, Hanning Zhang, Nan Jiang, and Tong Zhang. 2024.
\newblock An implementation of generative prm.
\newblock \url{https://github.com/RLHFlow/RLHF-Reward-Modeling}.

\bibitem[{Yang et~al.(2024)Yang, Yang, Zhang, Hui, Zheng, Yu, Li, Liu, Huang, Wei, Lin, Yang, Tu, Zhang, Yang, Yang, Zhou, Lin, Dang, Lu, Bao, Yang, Yu, Li, Xue, Zhang, Zhu, Men, Lin, Li, Xia, Ren, Ren, Fan, Su, Zhang, Wan, Liu, Cui, Zhang, and Qiu}]{DBLP:journals/corr/abs-2412-15115}
An~Yang, Baosong Yang, Beichen Zhang, Binyuan Hui, Bo~Zheng, Bowen Yu, Chengyuan Li, Dayiheng Liu, Fei Huang, Haoran Wei, Huan Lin, Jian Yang, Jianhong Tu, Jianwei Zhang, Jianxin Yang, Jiaxi Yang, Jingren Zhou, Junyang Lin, Kai Dang, Keming Lu, Keqin Bao, Kexin Yang, Le~Yu, Mei Li, Mingfeng Xue, Pei Zhang, Qin Zhu, Rui Men, Runji Lin, Tianhao Li, Tingyu Xia, Xingzhang Ren, Xuancheng Ren, Yang Fan, Yang Su, Yichang Zhang, Yu~Wan, Yuqiong Liu, Zeyu Cui, Zhenru Zhang, and Zihan Qiu. 2024.
\newblock \href {https://doi.org/10.48550/ARXIV.2412.15115} {Qwen2.5 technical report}.
\newblock \emph{CoRR}, abs/2412.15115.

\bibitem[{Yu et~al.(2024{\natexlab{a}})Yu, Gao, and Wang}]{DBLP:conf/naacl/YuGW24}
Fei Yu, Anningzhe Gao, and Benyou Wang. 2024{\natexlab{a}}.
\newblock \href {https://doi.org/10.18653/V1/2024.FINDINGS-NAACL.55} {Ovm, outcome-supervised value models for planning in mathematical reasoning}.
\newblock In \emph{Findings of the Association for Computational Linguistics: {NAACL} 2024, Mexico City, Mexico, June 16-21, 2024}, pages 858--875. Association for Computational Linguistics.

\bibitem[{Yu et~al.(2025)Yu, Li, and Wang}]{yu2025scaling}
Fei Yu, Yingru Li, and Benyou Wang. 2025.
\newblock Scaling flaws of verifier-guided search in mathematical reasoning.
\newblock \emph{arXiv preprint arXiv:2502.00271}.

\bibitem[{Yu et~al.(2024{\natexlab{b}})Yu, Jiang, Shi, Yu, Liu, Zhang, Kwok, Li, Weller, and Liu}]{DBLP:conf/iclr/YuJSYLZKLWL24}
Longhui Yu, Weisen Jiang, Han Shi, Jincheng Yu, Zhengying Liu, Yu~Zhang, James~T. Kwok, Zhenguo Li, Adrian Weller, and Weiyang Liu. 2024{\natexlab{b}}.
\newblock \href {https://openreview.net/forum?id=N8N0hgNDRt} {Metamath: Bootstrap your own mathematical questions for large language models}.
\newblock In \emph{The Twelfth International Conference on Learning Representations, {ICLR} 2024, Vienna, Austria, May 7-11, 2024}. OpenReview.net.

\bibitem[{Zelikman et~al.(2022)Zelikman, Wu, Mu, and Goodman}]{DBLP:conf/nips/ZelikmanWMG22}
Eric Zelikman, Yuhuai Wu, Jesse Mu, and Noah~D. Goodman. 2022.
\newblock \href {http://papers.nips.cc/paper\_files/paper/2022/hash/639a9a172c044fbb64175b5fad42e9a5-Abstract-Conference.html} {Star: Bootstrapping reasoning with reasoning}.
\newblock In \emph{Advances in Neural Information Processing Systems 35: Annual Conference on Neural Information Processing Systems 2022, NeurIPS 2022, New Orleans, LA, USA, November 28 - December 9, 2022}.

\bibitem[{Zhang et~al.(2024{\natexlab{a}})Zhang, Wu, Lei, Che, Li, Xie, Huang, Zhang, Pavone, Li, Ouyang, and Zhou}]{DBLP:journals/corr/abs-2410-02884}
Di~Zhang, Jianbo Wu, Jingdi Lei, Tong Che, Jiatong Li, Tong Xie, Xiaoshui Huang, Shufei Zhang, Marco Pavone, Yuqiang Li, Wanli Ouyang, and Dongzhan Zhou. 2024{\natexlab{a}}.
\newblock \href {https://doi.org/10.48550/ARXIV.2410.02884} {Llama-berry: Pairwise optimization for o1-like olympiad-level mathematical reasoning}.
\newblock \emph{CoRR}, abs/2410.02884.

\bibitem[{Zhang et~al.(2024{\natexlab{b}})Zhang, Wang, Diao, Lin, Pan, Dong, Zhang, Molchanov, and Zhang}]{DBLP:journals/corr/abs-2412-11006}
Hanning Zhang, Pengcheng Wang, Shizhe Diao, Yong Lin, Rui Pan, Hanze Dong, Dylan Zhang, Pavlo Molchanov, and Tong Zhang. 2024{\natexlab{b}}.
\newblock \href {https://doi.org/10.48550/ARXIV.2412.11006} {Entropy-regularized process reward model}.
\newblock \emph{CoRR}, abs/2412.11006.

\bibitem[{Zhang et~al.(2024{\natexlab{c}})Zhang, Hosseini, Bansal, Kazemi, Kumar, and Agarwal}]{DBLP:journals/corr/abs-2408-15240}
Lunjun Zhang, Arian Hosseini, Hritik Bansal, Mehran Kazemi, Aviral Kumar, and Rishabh Agarwal. 2024{\natexlab{c}}.
\newblock \href {https://doi.org/10.48550/ARXIV.2408.15240} {Generative verifiers: Reward modeling as next-token prediction}.
\newblock \emph{CoRR}, abs/2408.15240.

\bibitem[{Zhang et~al.(2025)Zhang, Zheng, Wu, Zhang, Lin, Yu, Liu, Zhou, and Lin}]{zhang2025lessons}
Zhenru Zhang, Chujie Zheng, Yangzhen Wu, Beichen Zhang, Runji Lin, Bowen Yu, Dayiheng Liu, Jingren Zhou, and Junyang Lin. 2025.
\newblock The lessons of developing process reward models in mathematical reasoning.
\newblock \emph{arXiv preprint arXiv:2501.07301}.

\bibitem[{Zhuang et~al.(2024)Zhuang, Chen, Yu, Mitra, Bursztyn, Rossi, Sarkhel, and Zhang}]{DBLP:conf/iclr/ZhuangC0MBRS024}
Yuchen Zhuang, Xiang Chen, Tong Yu, Saayan Mitra, Victor~S. Bursztyn, Ryan~A. Rossi, Somdeb Sarkhel, and Chao Zhang. 2024.
\newblock \href {https://openreview.net/forum?id=B6pQxqUcT8} {Toolchain*: Efficient action space navigation in large language models with a* search}.
\newblock In \emph{The Twelfth International Conference on Learning Representations, {ICLR} 2024, Vienna, Austria, May 7-11, 2024}. OpenReview.net.

\end{thebibliography}

\newpage
\appendix

\section{Experiment Details}
\subsection{Baselines.}\label{appendix:baseline}

\paragraph{Outcome Reward Model (ORM, \citealp{DBLP:journals/corr/abs-2110-14168}).} The vanilla ORM assigns a score to the entire solution as the final reward. We train ORMs through outcome supervision. Following \cite{DBLP:journals/corr/abs-2110-14168}, we replicate the binary correctness label  $r_t \in \{0, 1\}$  across the entire solution sequence. The reward head is then trained to predict reward scores for each token, enhancing robustness. 

\paragraph{Process Reward Model (PRM, \citealp{DBLP:conf/iclr/LightmanKBEBLLS24, DBLP:journals/corr/abs-2211-14275}).} The vanilla PRM assigns scores to each step along a solution path. For training stability , we place the reward label  $r_t$  at the last token of each step. In other words, for $t$-th step-level sequence, the label vector $\mathbf{y_t} = [0, 0, \dots, 0, r_t]$.

\paragraph{Math-shepherd PRM \cite{DBLP:conf/acl/WangLSXDLCWS24}.} Different from the vanilla PRM, Math-Shepherd PRM uses Monte-Carlo Estimation to annotate step labels. This estimation is essentially considered as training a value model \cite{zhang2025lessons}. In our experiments, we first sample $15$ solutions for each query. Then, for each intermediate step, we sample $8$ rollouts. We provide a detailed description of this method in Section~\ref{sec:value-label}.

\paragraph{ER-PRM \cite{DBLP:journals/corr/abs-2412-11006}.} Similar to Math-Shepherd PRM, ER-PRM integrates entropy-regularized step labels to train the supervision model. After Monte-Carlo sampling, ER-PRM calculates the label for the $t$ -th step according to the following equation:
\begin{equation}
\mathrm{label}(s_t) = \frac{1}{\eta}\ln \mathbb{E_{\tau^{-[\mathrm{t}]}\sim\pi}}e^{\eta y(\tau)} \nonumber
\end{equation}
where $\tau$ represents the complete rollout starting from the step $s_t$, $\pi$ represents the LLM generator, and $y(\cdot)$ represents the final correctness of the solution $\tau$ .

\subsection{BiRM Training Details}\label{appendix:birm-train}

In the BiRM training phase, we collect problems from the original GSM8K and MATH dataset . Then we use LLM generators to sample $15$ trajectories per query, resulting in a training set of approximately $225,000$ solutions for each base model. We annotate reward and value labels using the \texttt{Deepseek-V3} \cite{DBLP:journals/corr/abs-2412-19437} and Math-shepherd soft-label methods, respectively. We set training labels on the last token of each step, following \cite{DBLP:conf/acl/WangLSXDLCWS24}. The coefficient $c$ in Equation \ref{eq:birm-loss} is set to $1.0$.

\subsection{Evaluation Metrics}\label{appendix:eval}
At test-time, BiRM estimates both reward scores and value scores for partial solutions at the same time. We follow Equation \ref{eq:BiRM-eval} to calculate the final score. The coefficient $\beta$ for different base models on GSM8K, MATH-500, and Gaokao2023 are set to $\beta_\mathrm{Qwen2.5-3B} = \{1.0, 2.5, 2.0\}$, $\beta_\mathrm{Qwen2.5-7B} = \{1.5, 3.0, 3.5\}$, $\beta_\mathrm{Llama3.1-8B} = \{2.5, 1.0, 3.5\}$ respectively.

\paragraph{Best-of-N Sampling.} 
For a given question $q$, we sample multiple rollouts from the LLM, resulting in a candidate set of $N$ reasoning paths $\mathcal{T} = \{ \tau_1, \tau_2, \dots, \tau_N \}$. Subsequently, an additional supervision model $\mathcal{R}$ , such as PRM, is used to score each candidate path, yielding $\mathcal{R}(\tau_i)$, where $i \in \{1, 2, \dots, N\}$. The candidate with the highest score represents the most promising solution and is selected as the final output:
\begin{equation}
    \tau^* = \arg \max_{\tau \in \{\tau_1, \tau_2, \dots, \tau_N\}} \mathcal{R}(\tau) \nonumber
\end{equation}

\paragraph{Beam Search.}\label{appendix:beam-search} 
We present all search results from the main experiment in Table \ref{tab:app-beam-search-qwen-3b}, Table \ref{tab:app-beam-search-qwen-7b}, and Table \ref{tab:app-beam-search-llama-8b}. The procedure of the step-level beam search is as follows: We first set the total sampling size $K$ and beam size $b$ ( $K$  should be divisible by $b$) . In each round, we only expand one step forward. For a given query, we sample $K$ rollouts in the first round. Then, we use the supervision model $\mathcal{M}$ to re-rank these candidates and select the top $b$ rollouts for the next step. Starting from the second round, we expand $\frac{K}{b}$ trajectories for each candidate, getting $K$ candidates in total. We repeat the re-ranking process until a final answer is found or the maximum step count is reached. A detailed pseudocode is provided in \ref{algo:beam_search}.

\section{Step Label Annotation Details}
\subsection{Dataset preprocessing}
\begin{figure*}[t]
    \centering
    \includegraphics[width=0.96\textwidth]{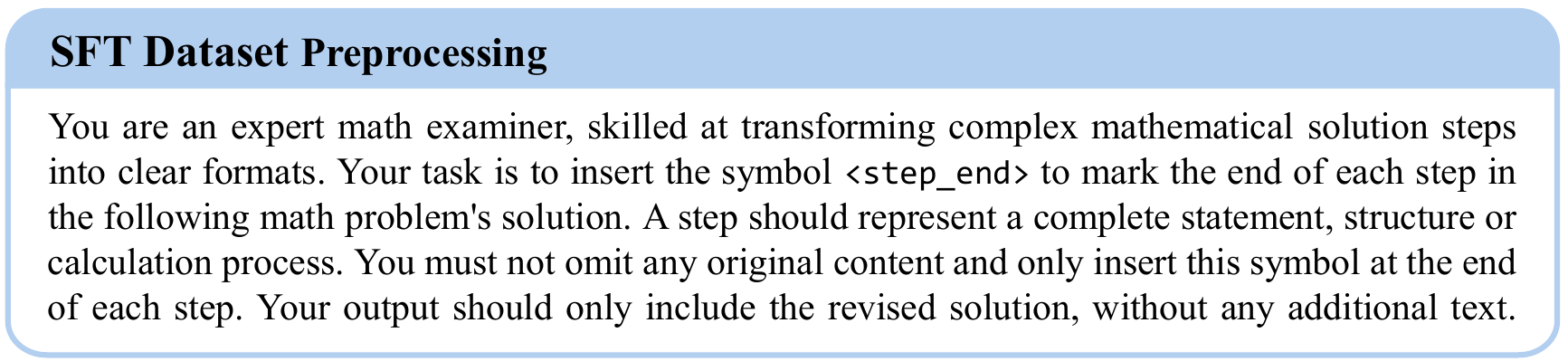}
    \caption{The prompt template for MetaMath dataset preprocessing.}
    \label{fig:app_prompt_1}
\end{figure*}

Before the SFT phase, we first preprocess the training data and restructure the delimiters at different levels of granularity. This is because we discover that original solution paths contain numerous meaningless text segments, which hinder the effective learning of process supervision models. Similar findings are reported by \cite{DBLP:conf/acl/Liao0LW024}. To address this, we utilize \texttt{Deepseek-V3} to clean the MATH subset in the MetaMath dataset, reannotate the delimiters, and ensure that each step is logically complete and meaningful. The prompt template for data preprocessing is shown in Figure \ref{fig:app_prompt_1}.

\subsection{Reward Label Annotaion}\label{appendix:reward-label}
\begin{figure*}[t]
    \centering
    \includegraphics[width=0.96\textwidth]{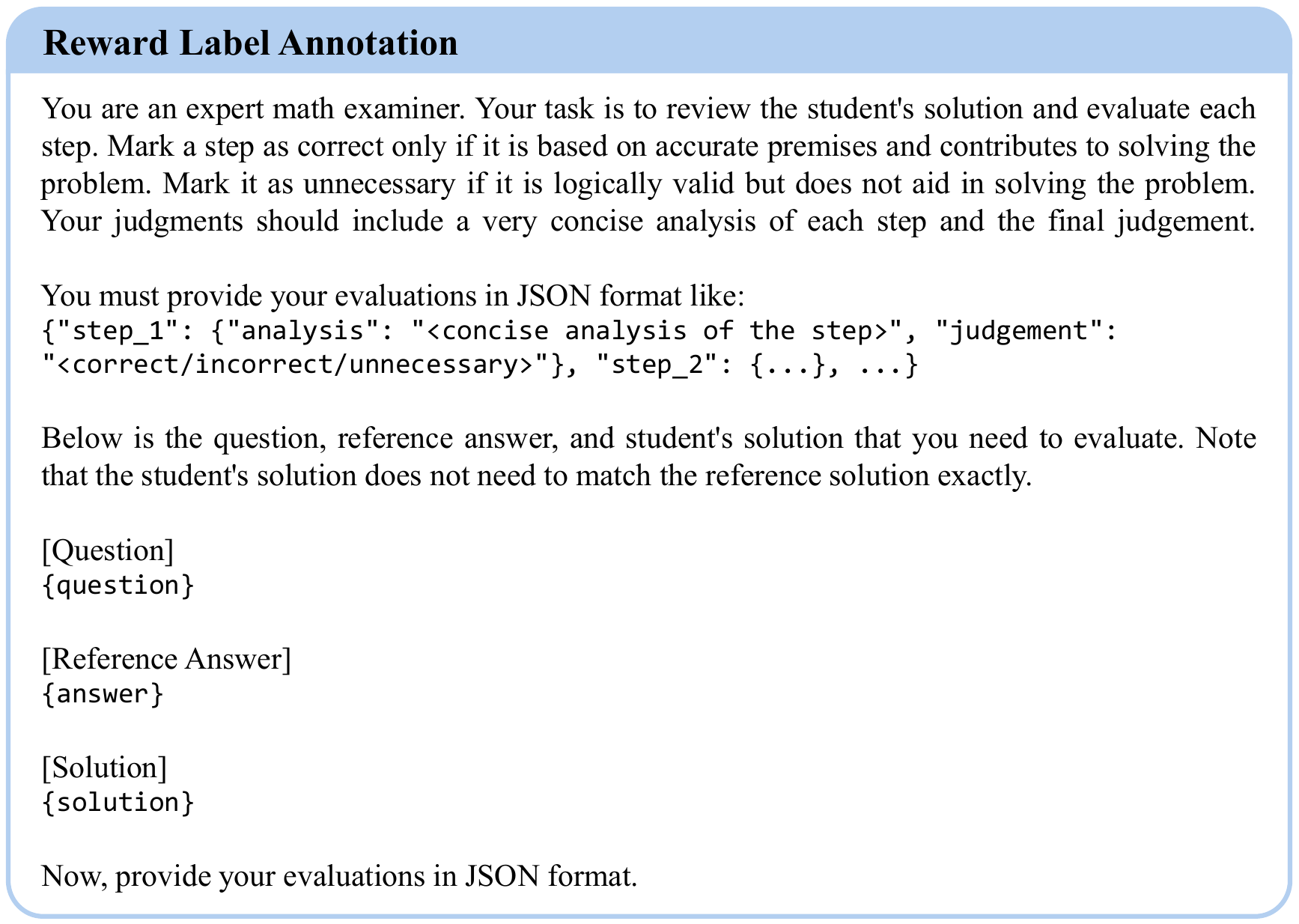}
    \caption{The prompt template for reward label annotaion.}
    \label{fig:app_prompt_2}
\end{figure*}

We also use \texttt{Deepseek-V3} to annotate the correctness of each step (i.e., reward label) in our experiments. The prompt template is provided in Figure~\ref{fig:app_prompt_2}.
\begin{algorithm}[ht]
\small
\caption{\label{algo:beam_search}Step-Level Beam Search}
\begin{algorithmic}[1]

\State $\textbf{Input:}$ Question $q$, Total Sampling Size $K$, Beam size $b$, Maximum step count $T$
\State $\textbf{Output:}$ Best solution path for $q$
\State $\textbf{Model:}$ Generator $\pi$ and BiRM $\mathcal{M}$

\Procedure{StepLevelBeamSearch}{$q, K, b$}
    \State Initialize partial solutions $\mathbb{T} \gets \{\}$
    \State Sample initial steps $\{\tau_1^1,\tau_2^1, \dots,\tau_K^1\}$
    \State Estimate scores $\{s_1^1, s_2^1, \cdots,s_K^1\}$ for each step
    \State Select top $b$ scored steps and add to $\mathbb{T}$
    \State $t \gets 1$
    \While{solutions in $\mathbb{T}$ are not complete and $t < T$}
        \State New candidate solutions $\mathbb{T}_{\text{new}} \gets \{\}$
        \State Scores $\mathcal{S} \gets \{\}$
        \For{each partial solution $\tau^{[1:t]}$ in $\mathbb{T}$}
            \For{$i = 1$ to $K/b$}
                \State $\tau^{[1:t+1]}_i \sim \pi(\tau^{[1:t]},q)$
                \State $s^{[1:t+1]}_i = \mathcal{M}(\tau^{[1:t+1]}_i,q)$
                \State $\mathbb{T}_{\text{new}} \gets \mathbb{T}_{\text{new}}+\tau^{[1:{t+1}]}_i$
                \State $\mathcal{S} \gets \mathcal{S}+s^{[1:t+1]}_i$
            \EndFor
        \EndFor
        \State $\mathbb{T}_{\text{new}} \gets $ top $b$ scored partial solutions in $\mathbb{T}_{\text{new}}$
        \State $\mathbb{T} \gets \mathbb{T}_{\text{new}}$
        \State $t \gets t+1$
    \EndWhile
    \State \textbf{return} solution with the highest score in $\mathbb{T}$
\EndProcedure

\end{algorithmic}
\end{algorithm}
\begin{table*}[t]
    \centering
    \resizebox{\textwidth}{!}{%
    \begin{tabular}{cc|ccc|ccc|ccc}
    \toprule
    \multirow{2}{*}{\textbf{Total Size}} & \multirow{2}{*}{\textbf{Beam Size}} & \multicolumn{3}{c|}{\textbf{GSM8K}} & \multicolumn{3}{c|}{\textbf{MATH-500}} & \multicolumn{3}{c}{\textbf{Gaokao2023}} \\
    \cmidrule(lr){3-11}
    & & OVM & PRM & BiRM & OVM & PRM & BiRM & OVM & PRM & BiRM \\
    \midrule
    
    \multirow{3}{*}{$K=4$} 
    & 4 & 81.50 ± 0.45 & 82.11 ± 0.28 & 81.96 ± 0.38 & 48.60 ± 0.16 & 48.13 ± 1.20 & 47.27 ± 0.96 & 33.85 ± 0.53 & 33.16 ± 0.74 & 33.07 ± 0.44 \\
    & 2 & 82.97 ± 0.14 & 81.53 ± 0.72 & 82.76 ± 0.28 & 48.27 ± 1.61 & 49.27 ± 0.81 & 50.07 ± 0.84 & 35.15 ± 1.80 & 34.55 ± 0.56 & 36.10 ± 0.76 \\
    & 1 & 80.82 ± 0.51 & 80.57 ± 0.77 & 81.93 ± 0.64 & 47.60 ± 0.59 & 47.80 ± 1.14 & 47.67 ± 0.57 & 35.58 ± 1.29 & 34.89 ± 1.56 & 32.47 ± 0.56 \\
    \midrule
    
    \multirow{4}{*}{$K=8$} 
    & 8 & 83.70 ± 0.73 & 83.65 ± 0.25 & 84.41 ± 0.35 & 49.33 ± 0.38 & 50.13 ± 1.00 & 50.07 ± 0.66 & 35.93 ± 1.17 & 34.63 ± 0.49 & 35.06 ± 1.39 \\
    & 4 & 84.61 ± 0.56 & 83.93 ± 0.16 & 85.11 ± 0.40 & 48.87 ± 0.68 & 50.87 ± 0.41 & 52.53 ± 0.90 & 36.10 ± 1.10 & 37.92 ± 0.76 & 37.92 ± 0.21 \\
    & 2 & 84.10 ± 0.36 & 83.17 ± 0.25 & 84.00 ± 0.39 & 50.07 ± 1.06 & 50.33 ± 0.94 & 50.67 ± 1.64 & 35.32 ± 1.10 & 37.58 ± 1.09 & 36.97 ± 1.07 \\
    & 1 & 83.27 ± 0.40 & 82.66 ± 1.05 & 82.99 ± 0.23 & 48.47 ± 2.03 & 49.67 ± 0.25 & 49.93 ± 0.41 & 33.42 ± 1.59 & 35.24 ± 0.32 & 35.06 ± 1.12 \\
    \midrule
    
    \multirow{6}{*}{$K=20$} 
    & 20 & 85.27 ± 0.04 & 85.65 ± 0.50 & 86.13 ± 0.11 & 52.13 ± 1.15 & 53.20 ± 0.59 & 53.33 ± 1.48 & 36.54 ± 0.44 & 35.67 ± 1.56 & 36.10 ± 1.85 \\
    & 10 & 86.73 ± 0.65 & 84.66 ± 0.34 & 86.91 ± 0.25 & 53.00 ± 0.16 & 54.27 ± 0.77 & 55.00 ± 0.65 & 37.66 ± 1.48 & 38.35 ± 1.24 & 37.23 ± 1.38 \\
    & 5 & 86.23 ± 0.28 & 84.86 ± 0.54 & 86.28 ± 0.33 & 52.20 ± 0.59 & 53.40 ± 0.85 & 54.27 ± 0.52 & 36.88 ± 1.06 & 37.49 ± 0.86 & 37.58 ± 0.24 \\
    & 4 & 86.20 ± 0.16 & 84.76 ± 0.22 & 85.04 ± 0.19 & 51.73 ± 0.66 & 51.80 ± 0.75 & 53.60 ± 0.75 & 37.49 ± 0.88 & 35.41 ± 1.00 & 39.05 ± 1.17 \\
    & 2 & 85.32 ± 0.25 & 84.74 ± 0.36 & 85.19 ± 0.19 & 49.00 ± 0.33 & 50.33 ± 1.00 & 51.80 ± 0.49 & 35.67 ± 0.44 & 35.84 ± 0.97 & 36.62 ± 0.85 \\
    & 1 & 83.60 ± 0.09 & 82.56 ± 0.74 & 84.23 ± 0.39 & 49.00 ± 0.91 & 50.67 ± 0.90 & 50.87 ± 0.84 & 34.29 ± 1.29 & 34.46 ± 1.41 & 37.06 ± 2.51 \\
    \midrule
    
    \multirow{3}{*}{$K=100$} 
    & 50 & 87.29 ± 0.22 & 85.87 ± 0.64 & 87.34 ± 0.22 & 52.87 ± 0.82 & 53.87 ± 0.19 & 55.13 ± 0.34 & 37.06 ± 0.74 & 37.40 ± 0.97 & 38.96 ± 0.92 \\
    & 25 & 87.54 ± 0.26 & 85.52 ± 0.80 & 87.64 ± 0.65 & 53.00 ± 1.50 & 53.20 ± 0.33 & 54.73 ± 0.75 & 38.10 ± 1.17 & 37.75 ± 1.09 & 38.18 ± 0.21 \\
    & 10 & 85.90 ± 0.33 & 84.51 ± 0.77 & 86.71 ± 0.37 & 51.27 ± 1.32 & 49.80 ± 0.57 & 53.40 ± 1.23 & 38.01 ± 1.05 & 37.92 ± 1.10 & 37.40 ± 1.85 \\
    \bottomrule
    \end{tabular}%
    }
    \caption{Qwen2.5-3B performance of beam search on GSM8K, MATH-500 and Gaokao2023.}
    \label{tab:app-beam-search-qwen-3b}
\end{table*}

\begin{table*}[t]
    \centering
    \resizebox{\textwidth}{!}{%
    \begin{tabular}{cc|ccc|ccc|ccc}
    \toprule
    \multirow{2}{*}{\textbf{Total Size}} & \multirow{2}{*}{\textbf{Beam Size}} & \multicolumn{3}{c|}{\textbf{GSM8K}} & \multicolumn{3}{c|}{\textbf{MATH-500}} & \multicolumn{3}{c}{\textbf{Gaokao2023}} \\
    \cmidrule(lr){3-11}
    & & OVM & PRM & BiRM & OVM & PRM & BiRM & OVM & PRM & BiRM \\
    \midrule
    
    \multirow{3}{*}{$K=4$} 
    & 4 & 86.10 ± 0.52 & 86.48 ± 0.43 & 87.04 ± 0.16 & 55.73 ± 1.52 & 53.73 ± 1.51 & 57.13 ± 1.15 & 40.09 ± 0.12 & 41.13 ± 1.24 & 43.90 ± 0.92 \\
    & 2 & 86.20 ± 0.53 & 86.00 ± 0.25 & 86.99 ± 0.13 & 55.53 ± 0.47 & 55.80 ± 1.72 & 55.87 ± 0.93 & 42.77 ± 0.24 & 42.68 ± 0.88 & 43.55 ± 0.61 \\
    & 1 & 85.65 ± 1.01 & 85.04 ± 0.50 & 86.76 ± 0.31 & 53.80 ± 1.77 & 54.93 ± 1.46 & 56.33 ± 1.04 & 41.47 ± 0.74 & 43.98 ± 0.12 & 44.50 ± 1.17 \\
    \midrule
    
    \multirow{4}{*}{$K=8$} 
    & 8 & 86.73 ± 0.62 & 88.12 ± 0.36 & 88.93 ± 0.49 & 58.27 ± 0.50 & 58.20 ± 0.71 & 58.13 ± 0.90 & 44.19 ± 0.76 & 44.24 ± 0.68 & 45.11 ± 0.68 \\
    & 4 & 88.63 ± 0.19 & 87.89 ± 0.73 & 89.36 ± 0.22 & 57.60 ± 0.85 & 59.07 ± 1.32 & 59.53 ± 1.24 & 43.72 ± 0.86 & 45.63 ± 1.21 & 46.84 ± 0.65 \\
    & 2 & 88.55 ± 0.37 & 87.45 ± 0.19 & 88.30 ± 0.53 & 57.00 ± 0.43 & 57.20 ± 1.56 & 58.67 ± 1.27 & 43.98 ± 1.41 & 44.68 ± 0.56 & 45.45 ± 0.97 \\
    & 1 & 87.57 ± 0.38 & 86.45 ± 0.40 & 87.47 ± 0.19 & 54.67 ± 1.09 & 57.27 ± 1.23 & 57.73 ± 1.32 & 44.24 ± 1.09 & 44.94 ± 1.27 & 43.38 ± 0.52 \\
    \midrule
    
    \multirow{6}{*}{$K=20$} 
    & 20 & 86.33 ± 0.38 & 88.65 ± 0.09 & 90.04 ± 0.58 & 59.07 ± 0.82 & 59.60 ± 0.59 & 60.33 ± 0.68 & 44.76 ± 1.38 & 45.89 ± 1.21 & 47.71 ± 0.44 \\
    & 10 & 90.40 ± 0.18 & 89.18 ± 0.42 & 90.40 ± 0.65 & 58.73 ± 1.16 & 61.53 ± 1.05 & 62.27 ± 1.09 & 45.19 ± 0.76 & 48.14 ± 0.74 & 48.23 ± 0.12 \\
    & 5 & 90.30 ± 0.12 & 88.98 ± 0.26 & 90.60 ± 0.28 & 57.53 ± 0.34 & 59.40 ± 0.49 & 60.73 ± 0.19 & 45.45 ± 0.64 & 47.36 ± 1.17 & 47.36 ± 1.22 \\
    & 4 & 89.56 ± 0.25 & 87.52 ± 0.62 & 89.94 ± 0.09 & 56.53 ± 0.90 & 58.40 ± 1.28 & 59.67 ± 0.34 & 45.02 ± 0.53 & 45.54 ± 0.88 & 48.31 ± 2.21 \\
    & 2 & 88.55 ± 0.06 & 88.05 ± 0.07 & 89.69 ± 0.34 & 56.93 ± 0.90 & 57.47 ± 1.11 & 58.87 ± 0.62 & 43.72 ± 1.07 & 44.59 ± 0.74 & 47.62 ± 0.96 \\
    & 1 & 87.79 ± 0.22 & 86.96 ± 0.47 & 88.07 ± 0.38 & 56.27 ± 0.34 & 57.73 ± 1.52 & 58.33 ± 0.77 & 42.68 ± 1.71 & 45.63 ± 0.74 & 45.80 ± 0.86 \\
    \midrule
    
    \multirow{3}{*}{$K=100$} 
    & 50 & 91.00 ± 0.22 & 88.32 ± 0.57 & 91.28 ± 0.12 & 60.13 ± 0.47 & 60.73 ± 0.34 & 62.53 ± 0.77 & 46.84 ± 0.44 & 48.31 ± 0.85 & 49.96 ± 0.32 \\
    & 25 & 91.18 ± 0.53 & 88.40 ± 0.21 & 91.66 ± 0.33 & 58.40 ± 1.23 & 59.27 ± 0.34 & 62.00 ± 0.98 & 46.32 ± 0.53 & 47.97 ± 0.12 & 47.62 ± 0.68 \\
    & 10 & 89.97 ± 0.25 & 88.15 ± 0.16 & 91.00 ± 0.09 & 57.47 ± 1.32 & 59.20 ± 1.31 & 61.20 ± 0.43 & 43.64 ± 0.42 & 46.93 ± 0.49 & 49.00 ± 0.32 \\
    \bottomrule
    \end{tabular}%
    }
    \caption{Qwen2.5-7B performance of beam search on GSM8K, MATH-500 and Gaokao2023.}
    \label{tab:app-beam-search-qwen-7b}
\end{table*}

\begin{table*}[t]
    \centering
    \resizebox{\textwidth}{!}{%
    \begin{tabular}{cc|ccc|ccc|ccc}
    \toprule
    \multirow{2}{*}{\textbf{Total Size}} & \multirow{2}{*}{\textbf{Beam Size}} & \multicolumn{3}{c|}{\textbf{GSM8K}} & \multicolumn{3}{c|}{\textbf{MATH-500}} & \multicolumn{3}{c}{\textbf{Gaokao2023}} \\
    \cmidrule(lr){3-11}
    & & OVM & PRM & BiRM & OVM & PRM & BiRM & OVM & PRM & BiRM \\
    \midrule
    
    \multirow{3}{*}{$K=4$} 
    & 4 & 71.44 ± 0.36 & 71.65 ± 0.33 & 71.37 ± 0.41 & 37.53 ± 0.66 & 38.67 ± 0.50 & 38.07 ± 0.68 & 23.81 ± 0.65 & 24.94 ± 0.21 & 23.29 ± 0.86 \\
    & 2 & 72.76 ± 0.41 & 71.11 ± 1.11 & 72.91 ± 0.80 & 38.53 ± 2.22 & 39.87 ± 0.68 & 40.73 ± 0.52 & 23.90 ± 1.10 & 25.11 ± 1.36 & 26.06 ± 0.68 \\
    & 1 & 70.74 ± 0.16 & 68.99 ± 0.67 & 71.57 ± 0.38 & 36.80 ± 0.75 & 39.60 ± 0.85 & 39.33 ± 0.98 & 23.20 ± 0.74 & 24.33 ± 0.74 & 24.59 ± 0.65 \\
    \midrule
    
    \multirow{4}{*}{$K=8$} 
    & 8 & 76.52 ± 0.45 & 75.92 ± 0.20 & 76.90 ± 0.53 & 39.93 ± 1.48 & 39.00 ± 0.59 & 41.27 ± 0.09 & 25.63 ± 1.07 & 25.71 ± 0.56 & 26.75 ± 0.56 \\
    & 4 & 77.36 ± 0.47 & 75.84 ± 0.54 & 78.32 ± 0.55 & 40.20 ± 0.49 & 40.13 ± 1.64 & 43.27 ± 0.57 & 25.19 ± 1.29 & 26.49 ± 1.53 & 27.45 ± 1.56 \\
    & 2 & 75.51 ± 0.49 & 73.79 ± 1.02 & 76.17 ± 0.04 & 39.07 ± 0.50 & 39.80 ± 1.85 & 41.47 ± 1.23 & 25.02 ± 0.96 & 26.58 ± 2.04 & 26.23 ± 1.18 \\
    & 1 & 74.00 ± 0.66 & 72.40 ± 0.57 & 74.60 ± 0.62 & 37.27 ± 1.64 & 40.13 ± 1.57 & 41.47 ± 0.77 & 23.72 ± 0.74 & 25.28 ± 1.59 & 25.80 ± 1.17 \\
    \midrule
    
    \multirow{6}{*}{$K=20$} 
    & 20 & 79.93 ± 0.22 & 79.23 ± 0.48 & 80.46 ± 0.29 & 41.53 ± 0.84 & 41.00 ± 0.75 & 44.13 ± 0.19 & 26.15 ± 0.74 & 25.71 ± 0.21 & 27.62 ± 0.44 \\
    & 10 & 81.40 ± 0.25 & 78.82 ± 0.22 & 81.73 ± 0.62 & 40.73 ± 0.68 & 41.60 ± 0.86 & 44.27 ± 0.34 & 25.97 ± 0.97 & 28.57 ± 1.18 & 29.18 ± 0.86 \\
    & 5 & 79.76 ± 0.37 & 76.90 ± 0.35 & 81.00 ± 0.20 & 40.80 ± 0.71 & 42.07 ± 1.09 & 43.93 ± 1.00 & 27.01 ± 0.37 & 28.14 ± 0.98 & 28.57 ± 0.73 \\
    & 4 & 79.56 ± 0.26 & 76.02 ± 0.36 & 80.16 ± 0.64 & 40.13 ± 1.55 & 42.00 ± 1.07 & 44.00 ± 0.43 & 24.33 ± 0.32 & 28.23 ± 0.12 & 26.93 ± 0.80 \\
    & 2 & 77.96 ± 0.60 & 75.39 ± 1.04 & 79.53 ± 1.08 & 39.07 ± 0.90 & 39.53 ± 0.50 & 42.40 ± 0.65 & 26.58 ± 1.41 & 26.75 ± 0.85 & 26.84 ± 0.12 \\
    & 1 & 76.27 ± 0.80 & 73.24 ± 1.02 & 78.17 ± 0.87 & 39.27 ± 0.82 & 40.00 ± 1.82 & 41.53 ± 1.36 & 25.54 ± 0.24 & 27.36 ± 1.44 & 27.27 ± 0.56 \\
    \midrule
    
    \multirow{3}{*}{$K=100$} 
    & 50 & 82.71 ± 0.11 & 80.34 ± 0.84 & 85.39 ± 0.52 & 41.07 ± 0.50 & 42.33 ± 0.66 & 46.13 ± 0.98 & 26.23 ± 2.02 & 29.61 ± 0.37 & 30.65 ± 0.37 \\
    & 25 & 82.71 ± 0.61 & 78.44 ± 0.43 & 84.53 ± 0.60 & 40.93 ± 1.32 & 42.00 ± 0.71 & 44.07 ± 0.68 & 25.37 ± 0.61 & 28.14 ± 0.86 & 29.00 ± 0.61 \\
    & 10 & 81.10 ± 0.46 & 77.81 ± 0.93 & 83.35 ± 0.70 & 39.87 ± 0.77 & 40.27 ± 1.06 & 45.00 ± 0.16 & 25.80 ± 0.32 & 27.19 ± 0.86 & 29.70 ± 0.80 \\
    \bottomrule
    \end{tabular}%
    }
    \caption{Llama3.1-8B performance of beam search on GSM8K, MATH-500 and Gaokao2023.}
    \label{tab:app-beam-search-llama-8b}
\end{table*}

\end{document}